\pdfoutput=1
\RequirePackage[T1]{fontenc}
\documentclass[12pt]{article}

\usepackage{multirow}

\usepackage{multirow}

\usepackage{tikz}
\usepackage{tikz-qtree}
\usepackage{geometry}

\usetikzlibrary{positioning, shapes, arrows.meta}

\usepackage[frozencache,cachedir=.]{minted}

\usepackage{tcolorbox}
\tcbuselibrary{minted,breakable,xparse,skins}

\definecolor{bg}{gray}{0.95}
\DeclareTCBListing{mintedbox}{O{}m!O{}}{%
  breakable=true,
  listing engine=minted,
  listing only,
  minted language=#2,
  minted style=default,
  minted options={%
    gobble=0,
    breaklines=true,
    breakafter=,,
    fontsize=\scriptsize,
    numbersep=8pt,
    breaksymbol=\ ,
    #1},
  boxsep=0pt,
  left skip=0pt,
  right skip=0pt,
  left=10pt,%
  right=10pt,
  top=3pt,
  bottom=3pt,
  arc=7pt,
  leftrule=0pt,
  rightrule=0pt,
  bottomrule=2pt,
  toprule=0pt,  %
  colback=bg,
  colframe=purple!70,
  enhanced,
  #3}

\usepackage{listings}
\newtcolorbox{shadedblockquote}{
  colback=gray!20,
  colframe=gray!60,
  boxrule=0pt,
  boxsep=0pt,
  left=10pt,
  right=10pt,
  sharp corners
}

\usepackage{caption}
\usepackage{subcaption}
\usepackage[utf8]{inputenc} %
\usepackage[T1]{fontenc}    %
\usepackage{url}            %
\usepackage{booktabs}       %
\usepackage{amsfonts}       %
\usepackage{nicefrac}       %
\usepackage{
microtype}      %
\usepackage{xcolor}         %
\usepackage{amsmath}
\usepackage{amssymb}
\usepackage{mathtools}
\usepackage{slashed}
\usepackage{braket}
\usepackage{enumitem}

\usepackage[colorlinks,citecolor=darkgreen,linkcolor=firebrick,urlcolor=firebrick]{hyperref}

\allowdisplaybreaks
\usepackage{graphicx}
\usepackage{tikz}
\usepackage{tikz-cd}
\usepackage{times}
\usepackage{courier}
\usepackage{bm}
\usepackage{physics}
\usepackage{xcolor}
\usepackage{natbib}
\usepackage{mdframed}
\usepackage{nicefrac}
\usepackage{booktabs}
\usepackage{lipsum}
\usepackage{titlesec}
\usepackage{wrapfig,lipsum,booktabs}
\usepackage{authblk}
\usepackage{blindtext}

\usepackage{algorithm}
\usepackage{algpseudocode}
\newcommand{\commentsymbol}{//}%
\algrenewcommand\algorithmiccomment[1]{\hfill {\footnotesize \commentsymbol{} #1}}

\usepackage{titletoc}
\usepackage{todonotes}
\usepackage{authblk}
\usepackage{setspace}
\usepackage{dsfont} %

\usepackage[nameinlink,capitalize,noabbrev]{cleveref}

\usepackage{CJK}

\usepackage{array}
\usepackage{makecell}

\def\({\left(}
\def\){\right)}
\def\[{\left[}
\def\]{\right]}

\usepackage{dashbox}
\usepackage{xcolor}
\usepackage{colortbl}
\definecolor{lightyellow}{rgb}{1.0, 0.95, 0.7}
\definecolor{Blue}{rgb}{0, 0, 0.8}
\definecolor{blue}{rgb}{0,0,1}
\definecolor{darkgreen}{rgb}{0,0.40,0}
\definecolor{firebrick}{rgb}{0.698,0.133,0.133}
\newcommand*{\red}[1]{\textcolor{red}{#1}}

\newcommand*{\blue}[1]{\textcolor{blue}{#1}}

\definecolor{colorA}{rgb}{1,0,0}
\definecolor{colorB}{rgb}{0,0.3,1}
\definecolor{colorC}{rgb}{0.9,0.8,0.2}
\definecolor{colorD}{rgb}{0,0.65,0}
\definecolor{lesslightgray}{rgb}{0.5,0.5,0.5}
\definecolor{light-gray}{gray}{0.95}

\def\P{\mathbb{P}}

\let\cite\citep 

\usepackage{amsthm}
\makeatletter
\def\th@remark{%
  \thm@headfont{\bfseries}%
  \normalfont %
  \thm@preskip\topsep \divide\thm@preskip\tw@
  \thm@postskip\z@ %
}
\makeatother

\theoremstyle{definition}

\tcolorboxenvironment{theorem}{
  breakable,
  colback=black!10,
  colframe=white,%
  width=\dimexpr\linewidth+10pt\relax,%
  enlarge left by=-5pt,%
  enlarge right by=-5pt,%
  boxsep=5pt,%
  boxrule=0pt,
  left=0pt,right=0pt,top=0pt,bottom=0pt,
  sharp corners,
  before skip=\topsep,
  after skip=\topsep
}

\tcolorboxenvironment{lemma}{
  breakable,
  colback=black!10,
  colframe=white,%
  width=\dimexpr\linewidth+10pt\relax,%
  enlarge left by=-5pt,%
  enlarge right by=-5pt,%
  boxsep=5pt,%
  boxrule=0pt,
  left=0pt,right=0pt,top=0pt,bottom=0pt,
  sharp corners,
  before skip=\topsep,
  after skip=\topsep
}

\tcolorboxenvironment{corollary}{
  breakable,
  colback=black!10,
  colframe=white,%
  width=\dimexpr\linewidth+10pt\relax,%
  enlarge left by=-5pt,%
  enlarge right by=-5pt,%
  boxsep=5pt,%
  boxrule=0pt,
  left=0pt,right=0pt,top=0pt,bottom=0pt,
  sharp corners,
  before skip=\topsep,
  after skip=\topsep
}

\theoremstyle{definition}

\tcolorboxenvironment{definition}{
  breakable,
  colback=black!10,
  colframe=white,%
  width=\dimexpr\linewidth+10pt\relax,%
  enlarge left by=-5pt,%
  enlarge right by=-5pt,%
  boxsep=5pt,%
  boxrule=0pt,
  left=0pt,right=0pt,top=0pt,bottom=0pt,
  sharp corners,
  before skip=\topsep,
  after skip=\topsep
}

\tcolorboxenvironment{assumption}{
  breakable,
  colback=black!10,
  colframe=white,%
  width=\dimexpr\linewidth+10pt\relax,%
  enlarge left by=-5pt,%
  enlarge right by=-5pt,%
  boxsep=5pt,%
  boxrule=0pt,
  left=0pt,right=0pt,top=0pt,bottom=0pt,
  sharp corners,
  before skip=\topsep,
  after skip=\topsep
}

\tcolorboxenvironment{problem}{
  breakable,
  colback=black!10,
  colframe=white,%
  width=\dimexpr\linewidth+10pt\relax,%
  enlarge left by=-5pt,%
  enlarge right by=-5pt,%
  boxsep=5pt,%
  boxrule=0pt,
  left=0pt,right=0pt,top=0pt,bottom=0pt,
  sharp corners,
  before skip=\topsep,
  after skip=\topsep
}

\crefname{theorem}{Theorem}{Theorems}
\crefname{proposition}{Proposition}{Propositions}
\crefname{lemma}{Lemma}{Lemmas}
\crefname{corollary}{Corollary}{Corollaries}
\crefname{definition}{Definition}{Definitions}
\crefname{assumption}{Assumption}{Assumptions}
\crefname{remark}{Remark}{Remarks}
\crefname{problem}{Problem}{Problems}
\crefname{property}{Property}{property}

\tcolorboxenvironment{hypothesis}{
  breakable,
  colback=black!10,
  colframe=white,%
  width=\dimexpr\linewidth+10pt\relax,%
  enlarge left by=-5pt,%
  enlarge right by=-5pt,%
  boxsep=5pt,%
  boxrule=0pt,
  left=0pt,right=0pt,top=0pt,bottom=0pt,
  sharp corners,
  before skip=\topsep,
  after skip=\topsep
}
\crefname{hypothesis}{Hypothesis}{Hypothesises}

\tcolorboxenvironment{fact}{
  breakable,
  colback=black!10,
  colframe=white,%
  width=\dimexpr\linewidth+10pt\relax,%
  enlarge left by=-5pt,%
  enlarge right by=-5pt,%
  boxsep=5pt,%
  boxrule=0pt,
  left=0pt,right=0pt,top=0pt,bottom=0pt,
  sharp corners,
  before skip=\topsep,
  after skip=\topsep
}
\crefname{fact}{Fact}{Facts}

\tcolorboxenvironment{example}{
  breakable,
  colback=black!10,
  colframe=white,%
  width=\dimexpr\linewidth+10pt\relax,%
  enlarge left by=-5pt,%
  enlarge right by=-5pt,%
  boxsep=5pt,%
  boxrule=0pt,
  left=0pt,right=0pt,top=0pt,bottom=0pt,
  sharp corners,
  before skip=\topsep,
  after skip=\topsep
}
\crefname{example}{Example}{Examples}

\tcolorboxenvironment{question}{
  breakable,
  colback=black!10,
  colframe=white,%
  width=\dimexpr\linewidth+10pt\relax,%
  enlarge left by=-5pt,%
  enlarge right by=-5pt,%
  boxsep=5pt,%
  boxrule=0pt,
  left=0pt,right=0pt,top=0pt,bottom=0pt,
  sharp corners,
  before skip=\topsep,
  after skip=\topsep
}
\crefname{question}{Question}{Questions}

\numberwithin{equation}{section}
\numberwithin{theorem}{section}
\numberwithin{proposition}{section}
\numberwithin{definition}{section}
\numberwithin{lemma}{section}
\numberwithin{assumption}{section}
\numberwithin{remark}{section}

\usepackage{lipsum}

\makeatletter
\let\save@mathaccent\mathaccent
\newcommand*\if@single[3]{%
    \setbox0\hbox{${\mathaccent"0362{#1}}^H$}%
    \setbox2\hbox{${\mathaccent"0362{\kern0pt#1}}^H$}%
    \ifdim\ht0=\ht2 #3\else #2\fi
}
\newcommand*\rel@kern[1]{\kern#1\dimexpr\macc@kerna}
\newcommand*\widebar[1]{\@ifnextchar^{{\wide@bar{#1}{0}}}{\wide@bar{#1}{1}}}
\newcommand*\wide@bar[2]{\if@single{#1}{\wide@bar@{#1}{#2}{1}}{\wide@bar@{#1}{#2}{2}}}
\newcommand*\wide@bar@[3]{%
    \begingroup
    \def\mathaccent##1##2{%
        \let\mathaccent\save@mathaccent
        \if#32 \let\macc@nucleus\first@char \fi
        \setbox\z@\hbox{$\macc@style{\macc@nucleus}_{}$}%
        \setbox\tw@\hbox{$\macc@style{\macc@nucleus}{}_{}$}%
        \dimen@\wd\tw@
        \advance\dimen@-\wd\z@
        \divide\dimen@ 3
        \@tempdima\wd\tw@
        \advance\@tempdima-\scriptspace
        \divide\@tempdima 10
        \advance\dimen@-\@tempdima
        \ifdim\dimen@>\z@ \dimen@0pt\fi
        \rel@kern{0.6}\kern-\dimen@
        \if#31
        \overline{\rel@kern{-0.6}\kern\dimen@\macc@nucleus\rel@kern{0.4}\kern\dimen@}%
        \advance\dimen@0.4\dimexpr\macc@kerna
        \let\final@kern#2%
        \ifdim\dimen@<\z@ \let\final@kern1\fi
        \if\final@kern1 \kern-\dimen@\fi
        \else
        \overline{\rel@kern{-0.6}\kern\dimen@#1}%
        \fi
    }%
    \macc@depth\@ne
    \let\math@bgroup\@empty \let\math@egroup\macc@set@skewchar
    \mathsurround\z@ \frozen@everymath{\mathgroup\macc@group\relax}%
    \macc@set@skewchar\relax
    \let\mathaccentV\macc@nested@a
    \if#31
    \macc@nested@a\relax111{#1}%
    \else
    \def\gobble@till@marker##1\endmarker{}%
    \futurelet\first@char\gobble@till@marker#1\endmarker
    \ifcat\noexpand\first@char A\else
    \def\first@char{}%
    \fi
    \macc@nested@a\relax111{\first@char}%
    \fi
    \endgroup
    }
\makeatother

\makeatletter
\newcommand*{\redefinesymbolwitharg}[1]{%
  \expandafter\let\csname ltx#1\expandafter\endcsname\csname #1\endcsname
  \@namedef{#1}{\@ifnextchar{^}{\@nameuse{#1@}}{\@nameuse{#1@}^{}}}%
  \expandafter\def\csname #1@\endcsname^##1##2{%
     \csname ltx#1\endcsname\ifx!##1!\else^{##1}\fi\mathopen{}\mathclose\bgroup\left(##2\aftergroup\egroup\right)
     }%
}
\makeatother
\redefinesymbolwitharg{sin}
\redefinesymbolwitharg{cos}

\titlespacing\section{0pt}{4pt plus 4pt minus 2pt}{-2pt plus 2pt minus 2pt}
\titlespacing\subsection{0pt}{2pt plus 4pt minus 2pt}{-2pt plus 2pt minus 2pt}
\titlespacing\subsubsection{0pt}{2pt plus 4pt minus 2pt}{-2pt plus 2pt minus 2pt}

\usepackage{titletoc}

\newcommand{\sys}{{GenoArmory}\xspace}

\def\Snospace~{\S{}}

\newcommand{\sref}[2]{\hyperref[#2]{#1 \ref{#2}}}

\title{GenoArmory: A Unified Evaluation Framework for Adversarial Attacks on Genomic Foundation Models
}

\newcommand*{\email}[1]{\footnote{\href{mailto:#1}{\texttt{#1}}}}

\begin{document}

\begin{titlepage}

\begin{flushright}
Last Update: \today
\end{flushright}

\begin{center}

{
\Large \bfseries %
\begin{spacing}{1.15} %
GenoArmory: A Unified Evaluation Framework for Adversarial Attacks on Genomic Foundation Models
\end{spacing}
}

Haozheng Luo$^{\dagger*}$\email{hluo@u.northwestern.edu}
\quad
Chenghao Qiu$^{\flat*}$\email{chenghaoqiu@tamu.edu}
\quad
Yimin Wang$^{\S}$\email{wyimin@umich.edu}
\quad
Shang Wu$^{\dagger}$\email{shangwu2028@u.northwestern.edu}
\quad
Jiahao Yu $^{\dagger}$\email{jiahao.yu@northwestern.edu} \\
\quad
Zhenyu Pan $^{\dagger}$\email{zhenyupan@u.northwestern.edu}
\quad
Weian Mao $^{\P}$\email{weian@mit.edu}
\quad
Haoyang Fang $^{\parallel}$\email{haoyangf@alumni.cmu.edu}
\quad
Hao Xu$^{\Diamond}$\email{haxu@bwh.harvard.edu}
\\
Han Liu$^{\dagger\ddag}$\email{hanliu@northwestern.edu}
\quad
Binghui Wang$^{\natural}$\email{bwang70@iit.edu}
\quad
Yan Chen$^{\dagger}$\email{ychen@northwestern.edu}\\

\def\thefootnote{*}
\footnotetext{These authors contributed equally to this work.}

{\small
\begin{tabular}{ll}

$^\dagger\;$Department of Computer Science, Northwestern University\\
$^\flat\;$Department of Computer Science and Engineering, Texas A\&M University\\
$^\natural\;$Department of Computer Science,
Illinois Institute of Technology\\
$^\ddag\;$Department of Statistics and Data Science, Northwestern University\\
$^\S\;$Department of Computer Science and Engineering, University of Michigan\\
$^\P\;$ 
 Department of Electrical Engineering and Computer Science, Massachusetts Institute of Technology\\
$^\parallel\;$ Department of Computer Science, Carnegie Mellon University\\
$^\Diamond\;$ Department of Medicine at Brigham and Women’s Hospital and Harvard Medical School\\
\end{tabular}}

\end{center}

\noindent
We propose the \textbf{first} unified adversarial attack benchmark for Genomic Foundation Models (GFMs), named \textbf{GenoArmory}. Unlike existing GFM benchmarks, GenoArmory offers the first comprehensive evaluation framework to systematically assess the vulnerability of GFMs to adversarial attacks. Methodologically, we evaluate the adversarial robustness of five state-of-the-art GFMs using four widely adopted attack algorithms and three defense strategies. Importantly, our benchmark provides an accessible and comprehensive framework to analyze GFM vulnerabilities with respect to model architecture, quantization schemes, and training datasets. 
Additionally, we introduce \textbf{GenoAdv}, a new adversarial sample dataset designed to improve GFM safety.
Empirically, classification models exhibit greater robustness to adversarial perturbations compared to generative models, highlighting the impact of task type on model vulnerability.
Moreover, adversarial attacks frequently target biologically significant genomic regions, suggesting that these models effectively capture meaningful sequence features.

\end{titlepage}

{
\setlength{\parskip}{0em}
\setcounter{tocdepth}{2}
\tableofcontents
}
\setcounter{footnote}{0}
\thispagestyle{empty}

\clearpage
\setcounter{page}{1}

\section{Introduction}
\label{sec:intro}

The advent of Genomic Foundation Models (GFMs) has revolutionized the analysis and generation of DNA and RNA sequences \citep{zhou2025genomeocean,zhou2024dnabert, zhou2024dnabert2, ye2024genomics, nguyen2024sequence, dalla2024nucleotide, nguyen2024hyenadna, dnabert2021ji}. 
These models, pre-trained on extensive genomic datasets, have demonstrated exceptional performance across a variety of genomics tasks, leading to widespread adoption in both research and industry.  For instance, GFMs have shown proficiency in generating high-quality DNA and RNA sequences \citep{zhou2025genomeocean, nguyen2024sequence} and in species classification tasks \citep{zhou2024dnabert2,dalla2024nucleotide,dnabert2021ji}. 
In the realm of medical diagnostics, GFMs contribute significantly by predicting gene pathogenicity  \citep{sayeed2024gene}  and assessing genome-wide variant effects \citep{benegas2023dna}. Their capabilities extend to functional genomics, aiding in promoter detection \citep{fishman2025gena} and transcription factor prediction \citep{fu2025foundation, kabir2024dna}, which are crucial for understanding gene regulation mechanisms.
GFMs also are instrumental in RNA secondary structure prediction \citep{yang2024omnigenome}, a critical aspect of understanding RNA function and interactions.

Despite the remarkable advancements, GFMs face significant challenges, particularly concerning their robustness and security. GFMs, which process structured, high-dimensional, and low-redundancy inputs like DNA sequences, are especially susceptible to adversarial attacks---even minor perturbations, such as single-nucleotide variations, can lead to substantial biological consequences.  For instance, recent studies 
\citep{montserrat2023adversarial} 
have demonstrated that DNA language models, including DNABERT-2 and the Nucleotide Transformer, are vulnerable to various adversarial strategies including nucleotide-level substitutions, codon-level modifications, and backtranslation-based transformations. 
Such attacks can significantly degrade model performance in tasks like antimicrobial resistance gene classification and promoter detection. 
Moreover, the generative capabilities of GFMs can be exploited by the attacker---it could manipulate models like GenomeOcean \citep{zhou2025genomeocean} to produce biologically nonsensical sequences, potentially leading to harmful application, even including the design of bioweapons \citep{peppin2024reality}.

\begin{figure}[ht]
    \centering
    \includegraphics[width=\linewidth]{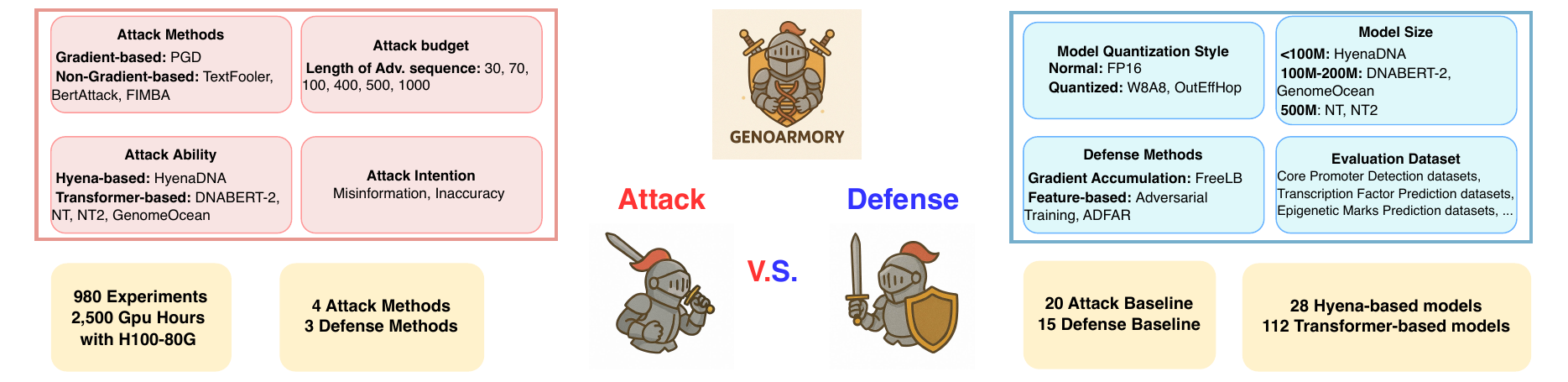}
    \caption{\textbf{An overview of benchmarking adversarial attacks on GFMs}}
    \label{fig:overview}
\end{figure}

{Given the significant safety concerns surrounding GFMs, there is a pressing need for robust defense mechanisms to ensure their reliability and security. 
However, the absence of benchmarks specifically designed to evaluate GFM safety has hindered the development of effective defense methods.
Existing efforts \citep{zhou2024dnabert2,liu2024genbench} primarily assess performance, without addressing safety aspects. 
This highlights the urgency of developing a new benchmark specifically designed to evaluate the safety of GFMs.}
To address this need, we introduce the GenoArmory benchmark, as shown in \cref{fig:overview}, designed to standardize best practices in the emerging field of adversarial attack and defense for DNA-based GFMs. 
GenoArmory is guided by core principles of transparency, reproducibility, and fairness in evaluating GFM robustness under both attack and defense scenarios. 
In this paper, we detail these guiding principles, describe the benchmark's components, report results across multiple attack and defense strategies on various GFMs, and share insights to inform robustness improvements.

\begin{figure}[htp]
    \centering
\includegraphics[width=1.0\linewidth]{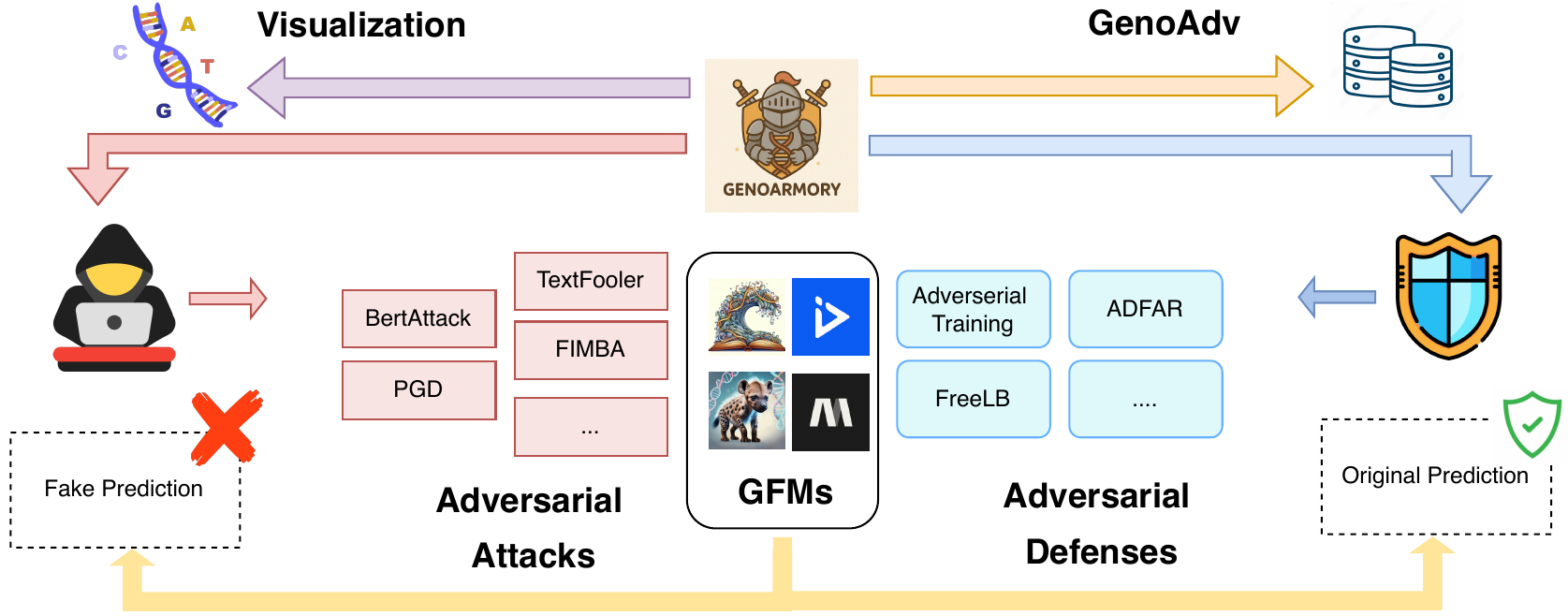}
    \caption{\textbf{GenoArmory Framework.} Our GenoArmory framework incorporates diverse adversarial attack and defense methods on GFMs. It also offers visualization tools to highlight important regions influencing model predictions and introduces a new adversarial dataset, \textbf{GenoAdv}.}
    \label{fig:pipeline}
\end{figure}

\textbf{Contributions:} We propose the GenoArmory framework (\cref{fig:pipeline}) to a comprehensively 
assess the robustness of GFMs against adversarial attacks. 
Our contributions include:
\begin{itemize}[leftmargin=*]
    \item \textbf{Pipeline for red-teaming GFMs.} We present a comprehensive evaluation pipeline to assess the robustness of DNA-based GFMs against adversarial attacks. 
    Specifically, our pipeline implements both gradient-based and gradient-free attack strategies across five different GFMs with standardized evaluation metrics.
    \item \textbf{Pipeline for testing and adding new defenses.} We implement three defense mechanisms and evaluate their effectiveness against adversarial attacks. 
    Additionally, we provide plug-and-play code to enable standardized evaluation of newly developed defense methods.
    \item \textbf{Repository of GFM adversarial attack artifacts.} We provide a repository of adversarial attack artifacts on GFMs, including adversarial examples and attack code, to facilitate reproducibility and further research in this area.
    \item \textbf{New adversarial sample dataset for GFMs.} We introduce a new dataset \textbf{GenoAdv}, composed of adversarial examples specifically generated to improve the robustness of GFMs. When used in training, GenoAdv yield a \textbf{34.71\%} Defense Success Rate, compared to training using only TextFooler samples.
    \item \textbf{Meaningful insights.} We provide a comprehensive analysis of GFM robustness under adversarial attacks, revealing the strengths and limitations of various models and defense strategies. 
    Additionally, we offer an in-depth discussion on how training methods and quantization settings impact the robustness of GFMs.
\end{itemize}

\section{Background}
\label{sec:related}
\textbf{Definition.} Given a genomic sequence $X = [x_1, x_2, \dots, x_n]$, where each nucleotide $x_i \in \{A, T, C, G\}$, a DNA model $f(\cdot) $, and a corresponding label $y$, our goal is to find an adversarial sequence $X'$ that satisfies:
\begin{equation*}
    f(X') \neq y \quad \text{subject to} \quad d(X, X') \leq \epsilon,
\end{equation*}
where $d(\cdot, \cdot)$ is a distance metric measuring the perturbation between the original and adversarial sequences, and $\epsilon$ controls the perturbation budget.

\textbf{Genomic Foundation Models.}
Recent advances in genomic foundation models (GFMs) \citep{liu2024genbench} establish two principal methodological paradigms: classification models and generative models. Within the classification paradigm, transformer-based approaches exhibit progressive technical refinements. Initial models, including DNABERT \citep{dnabert2021ji} and Nucleotide Transformer \citep{dalla2024nucleotide}, establish baseline performance through fixed k-mer tokenization strategies. DNABERT-2 \citep{zhou2024dnabert2} addresses these constraints by integrating byte-pair encoding (BPE) for tokenization and Attention with Linear Biases (ALiBi) for modeling longer sequences, which significantly enhances motif discovery capabilities. Building on this, DNABERT-S \citep{zhou2024dnabert} focuses on species differences in the embedding space. GERM \citep{luo2025fast} emerges as the first GFM specifically optimized for resource-constrained environments. By integrating an outlier-free architecture, GERM achieves both reliable quantization and fast adaptation. For long-range genomic dependency modeling, HyenaDNA \citep{nguyen2024hyenadna} replaces conventional attention mechanisms with Hyena operators, enabling efficient processing of ultra-long genomic sequences. 
Among generative models, GenomeOcean \citep{zhou2025genomeocean} represents a pioneer, trains on 220TB of genomic data, and demonstrates strong DNA sequence generation capabilities across diverse species domains. Meanwhile, Evo \citep{nguyen2024sequence} introduces a hybrid architecture that combines Hyena operators with sparse attention mechanisms capable of performing whole-genome modeling at single nucleotide resolution.

\textbf{Attack Methods.}
As shown in \cref{fig:adversarial_tree_styled}, adversarial attacks are broadly categorized into untargeted, targeted, and universal variants. Untargeted attacks \citep{FGSM, madry2017towards} aim to maximize model loss by perturbing inputs toward the gradient, while targeted attacks \citep{carlini2016towards, zhang2024lp} steer predictions toward specific classes by gradient. Universal attacks \citep{moosavi2017universal} generate input-agnostic perturbations that mislead models across entire data distributions. Numerous adversarial attack methods have been proposed in both NLP and CV, demonstrating their effectiveness in impacting model performance.
Only one work, FIMBA \citep{skovorodnikov2024fimba}, propose adversarial attacks in the genomic domain. FIMBA introduces a black-box, model-agnostic framework that perturbs key features identified via SHAP values to disrupt genomic models.

\textbf{Defense Methods.}
As shown in \cref{fig:adversarial_tree_styled}, defense strategies are broadly categorized into adversarial training, defensive distillation, adversarial sample detection, and regularization with certified robustness.
Adversarial training \citep{Zhu2020FreeLB,madry2017towards} enhances model robustness by iteratively injecting adversarial examples during training,
Another approach defensive distillation \citep{papernot2016distillation} trains student models on softened probability distributions from teacher models to smooth decision boundaries.
In contrast, adversarial sample \citep{Jin2024MAFD,zheng2023detecting,qi2021onion} detection identifies malicious inputs at inference time.
Regularization with certified robustness \citep{li-etal-2023-text,liu-etal-2022-flooding,ye-etal-2020-safer,jia-etal-2019-certified} reduces vulnerability through loss shaping.

\section{Main Features for \sys} 
\label{sec:method}
{Given the current landscape of GFMs, there exists no benchmark dedicated to evaluating their  reliability. Considering the significant safety concerns, we propose the \textbf{first} benchmark, \textbf{GenoArmory}, targeting adversarial attacks—one of the most critical threats to GFM security.
GenoArmory supports state-of-the-art attacks and defenses on GFMs,
as well as providing direct access to the corresponding adversarial attack artifacts.}
In particular, we prioritize the following aspects in our benchmark:
It will continuously update to incorporate emerging attacks and defenses from the literature.
Additionally, we aim to evolve the benchmark alongside the community to support newly developed methods.

\subsection{GenoAdv: A dataset of adversarial examples on GFMs}
\label{sub:dataset}
An important contribution of this work is the creation of an adversarial example dataset for GFMs, named \textbf{GenoAdv}.
This dataset comprises adversarial examples generated using multiple attack methods—BertAttack \citep{li2020bert}, TextFooler \citep{jin2020bert}, and FIMBA \citep{skovorodnikov2024fimba}—on various GFMs.
While prior studies \citep{li2020learning, zheng2020efficient, liu2019transferable} leverage transferable adversarial examples for training, the effectiveness of such transferability remains questionable.
To address this, we generate adversarial examples using diverse techniques to better capture model-specific vulnerabilities.
The GenoAdv dataset offers a comprehensive and diverse set of adversarial examples across different tasks and methods, providing users with a practical resource for rapid adversarial training to enhance model robustness.

\subsection{A repository of adversarial attacks artifacts}
\label{sub:attack}
\label{sub:artifacts}
A central component of the GenoArmory benchmark is our accessible repository of adversarial attack artifacts.  
Given the limited availability of GFM-specific adversarial attack method—FIMBA \citep{skovorodnikov2024fimba} being the only one to date—we adapt existing attack techniques from language and computer vision domains to GFMs.  
As a result, the GenoArmory artifact repository includes adversarial examples generated by BertAttack \citep{li2020bert}, TextFooler \citep{jin2020bert}, PGD \citep{pgd}, and FIMBA \citep{skovorodnikov2024fimba}.

\begin{mintedbox}[xleftmargin=0mm,autogobble]{python} 
from GenoArmory import GenoArmory
gen = GenoArmory(model="DNABERT-2-finetuned-H3", tokenizer="DNABERT-2-finetuned-H3")
gen.get_attack_metadata(method=TextFooler,model_name=dnabert)
\end{mintedbox}

\subsection{A pipeline for red-teaming GFMs}
Adversarial attacks on GFMs are challenging due to variations in tokenization, architecture, and datasets, leading to inconsistent results. To address this, we propose a standardized red-teaming pipeline that includes pre-trained GFMs, datasets, hyperparameters, and adversarial examples. The pipeline integrates five state-of-the-art models—DNABERT-2 \citep{zhou2024dnabert2}, Nucleotide Transformer (NT, NT2) \citep{dalla2024nucleotide}, GenomeOcean \citep{zhou2025genomeocean}, and HyenaDNA \citep{nguyen2024hyenadna}—along with 26 DNA-based classification datasets. It provides direct access to attack artifacts \cref{sub:artifacts} for standardized evaluation of adversarial robustness and supports user-defined attack methods, offering a flexible and extensible framework for evaluating model robustness.

\begin{mintedbox}[xleftmargin=0mm,autogobble]{python} 
import json
with open(params_file, "r") as f:
    kwargs = json.load(f)
gen.attack(attack_method='pgd', **kwargs)
\end{mintedbox}

\subsection{A pipeline for evaluating defenses against adversarial attacks}
\label{sub:defense}
In addition to efforts in developing new attack methods, researchers propose various defense strategies to counter adversarial threats.
Our benchmark provides a standardized pipeline for evaluating the effectiveness of these defenses against adversarial attacks.
Since no defense methods have been specifically designed for GFMs, we adapt existing state-of-the-arts from natural language and computer vision domains, i.e., adversarial training \citep{zheng2020efficient}, ADFAR \citep{adfar}, and FreeLB \citep{Zhu2020FreeLB}, as defense baselines for GFMs.
In our evaluation, we adopt existing attack methods as the base and assess the robustness of the defenses against adversarial examples generated by these attacks.

\begin{mintedbox}[xleftmargin=0mm,autogobble]{python} 
gen.defense(defense_method='freelb', **kwargs)
\end{mintedbox}

\subsection{Reproducible evaluation framework}
In addition to providing access to the attack artifacts and defense strategies,
we present a standardized evaluation framework, enabling users to benchmark robustness methods. 
The framework includes all essential components—data loading, model training and evaluation, and accuracy-based metrics.
A detailed discussion on reproducibility is provided in \cref{ap:reproduciblity}.

\subsection{A lightweight and easy-to-use implementation}
All implementations in our framework and pipelines are built on PyTorch and Huggingface Transformers \citep{wolf2019huggingface}. 
For defense evaluation, we employ the Hugging Face Trainer API to fine-tune the models. All resulting classification checkpoints will be publicly available on the Hugging Face Model Hub and can be easily downloaded and applied by researchers for further studies.

\subsection{A lightweight visulization framework}
\label{sub:visualization}
In our framework, we also introduce a visualization tool that enables users to explore how adversarial perturbations affect model predictions on input DNA sequences. 
Unlike language and computer vision domains—where explanations often rely on heuristic attribution or prediction maps—our approach leverages genomic knowledge to validate sequence-level changes with biological expectations.
Although there is a growing body of literature on explainable AI in the context of adversarial attacks \citep{moshe2024improving,devabhakthini2023analyzing,gipivskis2023impact,ozbulak2021investigating}, these works predominantly rely on saliency-based methods. 
In contrast, 
GFMs offer a promising path forward by grounding explanations in real-world biological data and leveraging bioinformatics for more interpretable and trustworthy insights.

\section{Evaluations of the Current Attacks and Defenses}
\label{sec:proof}

In this section, we conduct a series of experiments to assess the impact of adversarial attacks and defenses on the safety of GFMs. We use DNABERT-2 \citep{zhou2024dnabert2}, HyenaDNA \citep{nguyen2024hyenadna}, Nucleotide Transformer (NT) \citep{dalla2024nucleotide}, NT2, and GenomeOcean \citep{zhou2025genomeocean} as the target models.

\textbf{Models.} 
Following \citet{zhou2024dnabert2}, we use DNABERT-2, NT, NT2,  GenomeOcean, and HyenaDNA as target models which are trained specifically on DNA sequences. 
The first four are transformer-based models, whereas HyenaDNA utilizes a Hyena-based architecture. 
We finetune all models using the sequence classification technique, following \citet{zhou2024dnabert2}, and utilize the finetuned models 
as the targets to evaluate the adversarial attacks---we 
generate adversarial examples that are misclassified by the target models while indistinguishable from the original examples.

\begin{figure*}[!h]
    \centering
    \resizebox{0.8\textwidth}{!}{
    \includegraphics[width=\linewidth]{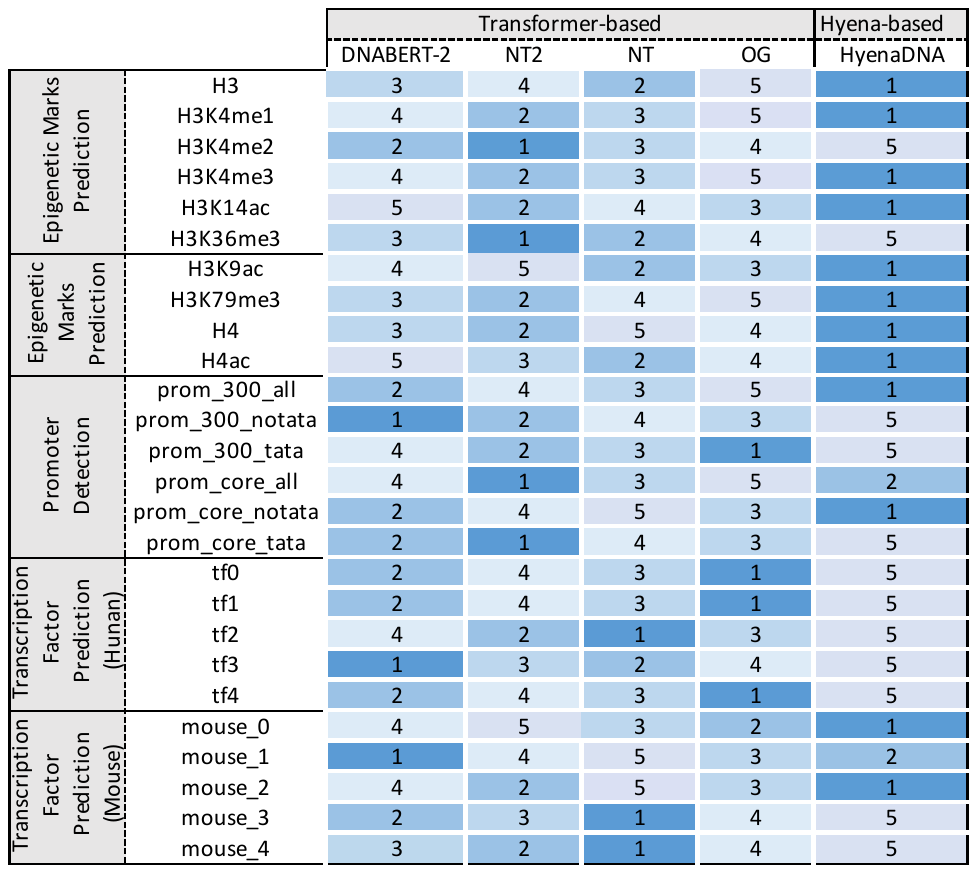}
    }
    \caption{\textbf{Performance of Adversarial Attacks on Different Model Architectures.} We assess the effectiveness of the evaluated adversarial attacks across diverse model architectures, including both transformer-based models (DNABERT-2, NT, NT2, GenomeOcean) and Hyena-based model (HyenaDNA). We use the Attack Success Rate (ASR) as the primary metric to evaluate the performance of the evaluated adversarial attacks. For each experiment, we rank the top five models based on their ASR, with ranks assigned from 1 to 5. A lower rank  indicates better robustness, while a higher rank reflects greater vulnerability to attacks.  Our results highlight how each model performs under attack, revealing differences in vulnerability and resilience across the architectures.}
    \label{fig:rank}
\end{figure*}

\textbf{Datasets.} 
We utilize 26 datasets covering 5 tasks and 4 species, as detailed in \citet{zhou2024dnabert2}. 
These datasets are specifically curated for genome sequence classification tasks, featuring input sequence lengths that range from 70 to 1000.

\textbf{Evaluation metrics.}
We evaluate the effectiveness of adversarial attacks using the Attack Success Rate (ASR) and assess defense strategies using the Defense Success Rate (DSR) as detailed in \cref{ap:metrics}. Accuracy is used as the core metric to quantify the impact of both attacks and defenses.

\begin{table}[ht]
  \centering
  \caption{\textbf{Adversarial Attack Performance of the Evaluated Method.} We conduct experiments to assess the effectiveness of the evaluated attack method against target models. The table presents a comparison of target model performance before and after applying the evaluated attack. We report Attack Success Rate (ASR) as the primary evaluation metric, with variance omitted as they are all $\le 2$\%.
  The final columns present the average Attack Success Rate (ASR) across all GFM models for each specific attack. The last row similarly shows the average ASR across all attacks for each specific GFM. Additionally, for each attack, individual ASR scores are ranked from \red{highest} to \blue{lowest}, with the rank displayed in brackets next to the score. }
  \label{tab:attack}
  \resizebox{\textwidth}{!}{
  \begin{tabular}{lcccccc}
    \toprule
    &  \multicolumn{4}{c}{Transformer-based} & Hyena-based \\
    \cmidrule(r){2-5} \cmidrule(l){6-6}
    Attack  & DNABERT-2 & NT & NT2 & GenomeOcean & HyenaDNA & Avg\\
    \midrule
    BertAttack & 96.23\%(\blue{5}) & 99.87\%(\red{1}) & 99.56\%(4) & 99.57\%(3) & 99.75\%(2) & 99.00\% \\
    TextFooler & 92.37\%(4) & 96.69\%(2) & 96.56\%(3) & 99.54\%(\red{1}) & 88.45\%(\blue{5}) &  94.72\%\\
    PGD        & 38.28\%(2) & 38.23\%(3) & 34.41\%(\blue{5}) & 36.57\%(4) & 47.94\%(\red{1}) & 39.09\%\\
    FIMBA      & 39.94\%(2) & 37.66\%(3) & 36.50\%(4) & 41.06\%(\red{1}) & 30.35\%(\blue{5}) & 37.10\% \\
    \cline{1-6}
    Attack ASR & 66.71\% (3.25) & 68.11\% (\red{2.25}) & 66.76\% (\blue{4}) & 69.19\% (\red{2.25}) & 66.62\% (3.25) & \\
    \bottomrule
  \end{tabular}
  }
\end{table}

\subsection{Evaluating adversarial attacks on GFMs}
\label{sub:attack2}
We utilize the same datasets and models as described in \cref{sub:attack} to ensure consistency in our evaluation.
We conduct each evaluation three times with different random seeds and present the average and standard deviation for each metric.

\textbf{Baseline attack artifacts.}
We test four baseline attack methods—BertAttack \citep{li2020bert}, TextFooler \citep{jin2020bert}, PGD \citep{pgd}, and FIMBA \citep{skovorodnikov2024fimba}—to assess their effectiveness in generating adversarial examples. Experiments are conducted on 5 GFMs, covering both transformer-based and Hyena-based architectures, with implementation details provided in \cref{ap:implementation}. Attack performance is primarily measured using ASR, and methods are ranked based on their average ASR across all datasets.

\textbf{Results.} In \cref{fig:rank,tab:attack}, our results highlight the effectiveness of the evaluated attacks in generating adversarial examples that are misclassified by target models. Further results are included in \cref{tab:addational_attack}.
We have below observations. 
\begin{itemize}[leftmargin=*]
\item GenomeOcean exhibits greater susceptibility to adversarial attacks than classification models (DNABERT-2, NT2), as evidenced by higher ASR and ranks across all GFMs. 
This observation aligns with the findings in \citet{ebrahimi2018,wang2023on}.

\item NT2 demonstrates the highest robustness, indicated by its lowest average rank, potentially due to its use of BPE tokenization. 
GFMs employing BPE tokenization (DNABERT-2, NT2) appear to be more robust than those using k-mer tokenization (NT). 
BPE's subword structure allows for partial token retention despite alterations, hindering significant semantic or biological shifts. 
Interestingly, while NT2's average ASR is higher than HyenaDNA's (the lowest overall), its ASR rank is lower. 
In contrast, NT shares the highest ASR rank with GenomeOcean but has a lower ASR. 
The discrepancy stems from NT consistently achieving high ASR across all attacks, while GenomeOcean performs best on TextFooler and FIMBA but poorly on BertAttack and PGD. 

\item BertAttack yields the highest average ASR across GFMs, while FIMBA, the only genome-specific attack, shows the lowest, indicating limited effectiveness. This ineffectiveness may be due to constraints in the released FIMBA code \footnote{https://github.com/HeorhiiS/fimba-attack} and evaluation setup in \citet{skovorodnikov2024fimba}. 
However, traditional NLP-based adversarial attacks such as BertAttack and TextFooler already achieve a high ASR in these models. 
This underscores the importance of developing defense mechanisms tailored for GFM tasks to ensure their safety.
\end{itemize}

\subsection{Evaluating adversarial defenses}
\label{sub:defense2}
Each experiment is repeated three times with different random seeds on the same datasets and models, and we report the mean and standard deviation of each evaluation metric.

\textbf{Baseline defenses.}
We assess 
the robustness of five GFM models against adversarial attacks using three defense baselines: adversarial training \citep{zheng2020efficient} (employing TextFooler for data augmentation), FreeLB \citep{Zhu2020FreeLB}, and ADFAR \citep{adfar}. 
Defenses were evaluated against BertAttack, TextFooler, and PGD attacks, with the DSR as the primary robustness metric. Further results are included in \cref{tab:addational_defense}.

\textbf{Results.}
As shown in \cref{tab:defense}, we have below observations: 
\begin{itemize}[leftmargin=*]
\item ADFAR achieves the highest overall DSR, significantly outperforming other defenses against BertAttack and TextFooler. 
However, ADFAR performs poorly against the PGD attack. 

\item FreeLB obtains better DSR against PGD, possibly due to it smooths the adversarial loss
during training, which somewhat improves robustness.

\item AT is less effective than ADFAR and FreeLB against BertAttack and TextFooler, although AT performs comparably to FreeLB against PGD attacks. 

 \item While the model architecture does not significantly affect overall defense performance, specific models show distinct advantages, e.g., DNABERT-2 and NT2 show a greater defense improvement against BertAttack, while HyenaDNA demonstrates a better defense against TextFooler and PGD.
\end{itemize}

\begin{table}[!t]
  \centering
  \caption{\textbf{Defense Performance Under Adversarial Attacks.} We conducted experiments to evaluate the performance of a defense method against adversarial attacks. 
  The table compares the performance of target models, both with and without the evaluated defense, under BertAttack, TextFooler, and PGD attacks. The Defense Success Rate (DSR) is used as the primary evaluation metric, with variance omitted as they are all $\le 2$\%. The best DSR values are highlighted in bold. In the table, \textbf{AT} denotes traditional adversarial training.
  We observe that ADFAR is the most effective defense based on DSR, particularly against BertAttack and TextFooler.
  }
  \label{tab:defense}
  \resizebox{0.9\textwidth}{!}{
  \begin{tabular}{lcccccc}
    \toprule
    & & \multicolumn{4}{c}{Transformer-based} & Hyena-based \\
    \cmidrule(r){3-6} \cmidrule(l){7-7}
    Attack Method & Defense  & DNABERT-2 & NT & NT2 & GenomeOcean & HyenaDNA\\
    \midrule
    \multirow{4}{*}{BertAttack}&N/A & 3.77\% & 0.13\% & 0.44\% & 0.43\% & 0.25\% \\
    &AT & 4.06\% & 0.21\% & 0.46\% & 0.60\% & 0.81\%\\
    &FreeLB & 4.34\% & 0.67\% & 0.71\% & \textbf{2.94\%} & 1.12\% \\
    &ADFAR & \textbf{21.84\%} & \textbf{4.95\%} &\textbf{ 6.96\%} & 1.18\% & \textbf{1.50\%}\\
    \midrule
    \multirow{4}{*}{PGD} &N/A & 61.73\% & 61.77\% & 65.59\% & 63.43\% & 52.06\%\\
    &AT &  \textbf{64.92\%} & 79.10\% & 82.02\% & \textbf{66.14\%} & 85.67\% \\
    &FreeLB & 64.07\% & \textbf{79.38\%} & \textbf{88.53\%} & 65.96\% & \textbf{86.99\%} \\
    &ADFAR & 63.48\% & 63.44\% & 72.89\% &  65.87\% & 83.74\% \\
    \midrule
    \multirow{4}{*}{TextFooler} &N/A & 7.63\% & 3.31\% & 3.44\% & 0.46\% & 11.55\% \\
    &AT &  20.97\% & 42.88\% & 18.95\% & 18.51\% & \textbf{84.19\%} \\
    &FreeLB & 18.39\% & 42.94\% & 18.16\% & 17.33\% & 69.56\%\\
    &ADFAR & \textbf{32.88\%} & \textbf{67.07\% }& \textbf{22.00\%} & \textbf{46.18\%} & 80.82\%\\
    \bottomrule
  \end{tabular}
  }
\end{table}

\begin{figure*}[!t]
    \centering
    \includegraphics[width=\linewidth]{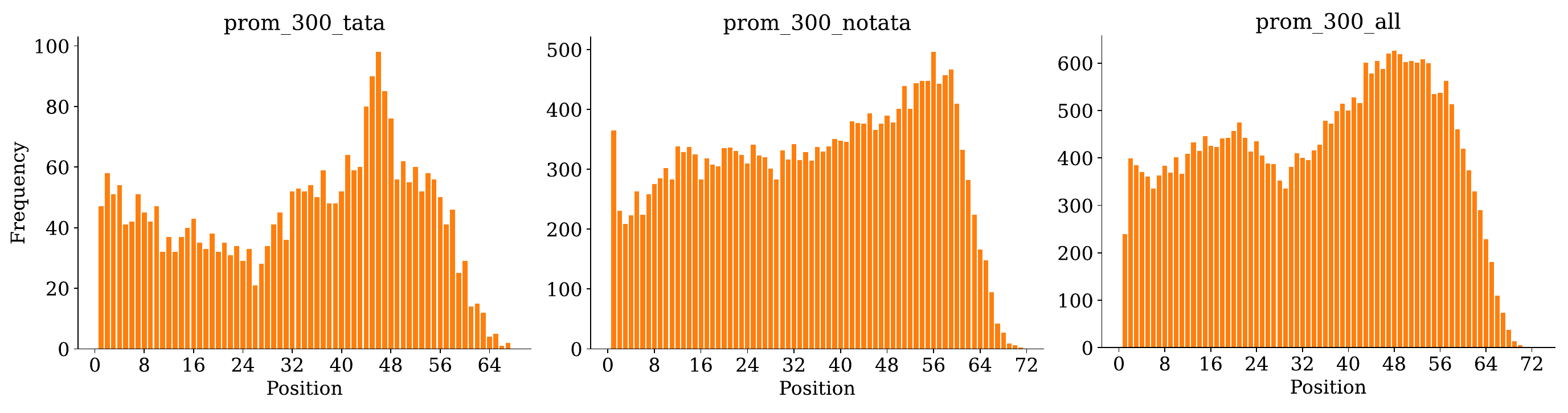}
    \caption{\textbf{Examples of the visualization of GFMs with adversarial attacks.} 
     We present the results of the three tasks of the DNABERT-2 model under BertAttack. 
    All subsequence changes occur at the subword tokenizer level using Byte Pair Encoding (BPE) \citep{sennrich2015neural}. 
    The visualization highlights which parts of the sequence are most significant for the model's classification performance. 
    Specifically, we present the frequency with which the adversarial attack modifies the sequence. 
    A higher frequency suggests that the subsequence plays a more important role in the model’s classification decisions.
    }
    \label{fig:visual}
\end{figure*}

\subsection{Visualization of adversarial attacks}
\label{sub:visual}
In this experiment, we visualize adversarial attacks on target models with our framework.
We utilize BertAttack to generate adversarial examples and visualize the results using the DNABERT-2 model.
The visualization highlights the subsequences that are most significant for the model's classification performance, specifically focusing on the frequency with which the adversarial attack modifies the sequence.
We present the frequency of subsequence changes at the subword tokenizer level using Byte Pair Encoding (BPE).
As shown in \cref{fig:visual}, the visualization is generated by analyzing the frequency of subsequence changes across all datasets and models, providing insight into the most critical subsequences for the model's classification performance.

\subsection{Performance of model augmented with GenoAdv dataset}
In order to show the effectiveness of the GenoAdv dataset, we conduct experiments to evaluate the performance of the model augmented with the GenoAdv dataset.
We use BertAttack, TextFooler, and PGD to evaluate the DSR on 5 GFMs. In our experiment, we perform traditional adversarial training with TextFooler-augmented data as a baseline, and compare it to the same training approach using the GenoAdv dataset.
We conduct each evaluation three times with different random seeds and present the average and standard deviation for each metric.

\textbf{Results:} As shown in \cref{tab:augmented}, adversarial training with GenoAdv data yields stronger robustness against adversarial attacks compared to training with only TextFooler-augmented samples in most cases. This suggests that the GenoAdv dataset offers valuable augmentation data to mitigate the vulnerability of GFMs. Specifically, using GenoAdv data to do data augmentation leads to a performance improvement of 34.71\% over TextFooler. Further results are included in \cref{tab:addational_augmented}.

\begin{table}[!t]
\centering
\caption{\textbf{Defense Performance Augmented with the GenoAdv Dataset.} 
We conduct experiments to evaluate the performance of a model augmented with the GenoAdv dataset against adversarial attacks.
The table compares the performance of the target models, both with and without the GenoAdv dataset augmentation, under BertAttack, TextFooler, and PGD attacks.
We report ASR as the primary evaluation metric, with variance omitted as they are all $\le 2$\%. The best results are highlighted in bold. In the table, \textbf{AT} denotes traditional adversarial training.
We observe that GenoAdv samples are more effective than TextFooler samples under traditional adversarial training methods. }
\label{tab:augmented}
\resizebox{0.9\textwidth}{!}{
  \begin{tabular}{lcccccc}
    \toprule
    & & \multicolumn{4}{c}{Transformer-based} & Hyena-based \\
    \cmidrule(r){3-6} \cmidrule(l){7-7}
    Attack Method & Defense  & DNABERT-2 & NT & NT2 & GenomeOcean & HyenaDNA\\
    \midrule
    \multirow{3}{*}{BertAttack}&N/A & 3.77\% & 0.13\% & 0.44\% & 0.43\% & 0.25\% \\
    &AT & 4.06\% & 0.21\% & 0.46\% & 0.60\% & 0.81\%\\
    &GenoAdv & \textbf{5.17\%} & \textbf{0.69\%} & \textbf{0.59\%} & \textbf{0.73\%} & \textbf{5.23\%} \\
    \midrule
     \multirow{3}{*}{PGD} &N/A & 61.73\% & 61.77\% & 65.59\% & 63.43\% & 52.06\%\\
      &AT &  64.92\% & 79.10\% & \textbf{82.02\%} & 66.14\% & \textbf{85.67\%} \\
    &GenoAdv &  \textbf{69.32\%} & \textbf{79.31\%} & 75.57\% & \textbf{67.10\%} & 84.52\% \\
    \midrule
     \multirow{3}{*}{TextFooler} &N/A & 7.63\% & 3.31\% & 3.44\% & 0.46\% & 11.55\% \\
     &AT &  20.97\% & 42.88\% & 18.95\% & 18.51\% & \textbf{84.19\%} \\
    &GenoAdv &  \textbf{22.19\%} & \textbf{44.05\%} & \textbf{20.56\%} & \textbf{19.45\%} & 81.99\% \\
    \bottomrule
  \end{tabular}
}
\end{table}

\subsection{Quantization influence on adversarial attacks}
\label{app:quantized}
To evaluate the influence of quantization on evaluated attacks, we conduct experiments on quantized versions of target models. 
Inside those quantization methods, some of them are based on the traditional quantization methods, such as uniform quantization, and some of them are based on the outluer-removal quantization methods, such as OutEffHop \citep{hu2024outlier}.
Following the quantization setup in \citet{luo2025fast} and \citet{wu2025sparq}, we evaluate the performance of the attacks on quantized models with 8-bit weights and 8-bit activations (W8A8), comparing them to the original models to analyze the impact of quantization on attack detectability.

\textbf{Results.} In \cref{tab:quantization}, our results highlight the effectiveness of quantization in improving the robustness of target models against adversarial attacks. Specifically, we observe that the evaluated attacks achieve a lower ASR on quantized models compared to the original models, indicating that quantization strengthens the defenses against these attacks. Additionally, the outlier-free quantization method also reduces the ASR of the evaluated attacks.
This outcome suggests that quantization can improve model robustness against adversarial attacks. One possible explanation is that quantization introduces "flat regions" in the loss landscape, which diminishes the model's sensitivity to small perturbations. This observation aligns with the findings reported in \citet{lin2018defensive}.

However, we find that the OutEffHop quantization method results in a higher ASR compared to traditional quantization methods, indicating that outlier-removal quantization can compromise the robustness of target models against adversarial attacks. 
A possible reason is that OutEffHop removes attention outliers which improves quantization, but also eliminates "flat regions" in the loss landscape that are important for robustness in traditional quantization methods.
We also find that quantization significantly impacts DNABERT-2 models, but has minimal effect on NT1 models, suggesting model-specific robustness gains. 
Notably, TextFooler is more affected by quantization than BERT-Attack, likely due to its dependence on precise word importance scores and synonym substitutions, which are disrupted by quantization-induced shifts in decision boundaries.

\begin{figure}[!tbp]
\centering 
\begin{minipage}{0.59\textwidth}
\captionsetup{type=table}
\caption{\textbf{Performance of the evaluated attacks on quantized models.
} We perform experiments to assess how quantization affects the effectiveness of adversarial attacks on target models. The table compares model performance before and after quantization under BertAttack and TextFooler attacks. Attack Success Rate (ASR) serves as the primary evaluation metric, with variance omitted as they are all $\le 2$\%. The best results are highlighted in bold.
}
\label{tab:quantization}
\end{minipage}
\hfill
\begin{minipage}{0.39\textwidth}
\resizebox{\textwidth}{!}{
\begin{tabular}{lccc}
\toprule
\textbf{Attack Method} & \textbf{Model} & \textbf{Quantized Method} & \textbf{ASR ($\downarrow$)} \\
\midrule
\multirow{6}{*}{BertAttack} & \multirow{3}{*}{DNABERT-2} & - & 96.23 \\
 & & Vanilla & \textbf{59.46} \\
& & OutEffHop & 64.71 \\
\cline{2-4}
 & \multirow{3}{*}{NT1} & - & 99.87 \\
 & & Vanilla & \textbf{99.37} \\
& & OutEffHop & 99.42 \\
\midrule
\multirow{6}{*}{TextFooler} & \multirow{3}{*}{DNABERT-2} & - & 92.37 \\
 & & Vanilla & \textbf{19.90} \\
& & OutEffHop &21.34\\
\cline{2-4}
 & \multirow{3}{*}{NT1} & - & 98.23 \\
 & & Vanilla & \textbf{66.57} \\
& & OutEffHop & 68.53 \\
\bottomrule
\end{tabular}
}
\end{minipage}
\end{figure}

\section{Discussion and Conclusion }
\label{sec:conclusion}We introduce GenoArmory, the first unified adversarial attack benchmark for DNA-based Genomic Foundation Models (GFMs). Our benchmark offers an accessible, reproducible, and comprehensive framework, enabling users to confidently evaluate and compare adversarial robustness in GFMs. Also, to encourage broad participation, we do not restrict the architectures of target models. Instead, GenoArmory offers a standardised framework for evaluating adversarial attacks and defenses.
Methodologically, compared to adversarial attack benchmarks in language and computer vision \citep{zheng2023blackboxbench,croce2020robustbench,dong2020benchmarking}, GenoArmory includes visualization tools that facilitate deeper insights into the evaluated attacks—leveraging the fact that GFM data is inherently structured and scientifically meaningful.

\clearpage

\appendix
\label{sec:append}
\part*{Appendix}
{
\setlength{\parskip}{-0em}
\startcontents[sections]
\printcontents[sections]{ }{1}{}
}

{
\setlength{\parskip}{-0em}
\startcontents[sections]
\printcontents[sections]{ }{1}{}
}

\section{Open Science}
\label{ap:reproduce}
We release the code, pretrained checkpoints, and datasets used in our work. The code is available at \href{https://github.com/MAGICS-LAB/GenoArmory}{this GitHub repository}, and the pretrained checkpoints are hosted on \href{https://huggingface.co/collections/magicslabnu/gfm-67f4d4a9327ee4acdcb3806b}{HuggingFace}. The GenoAdv dataset is hosted on Hugging Face \href{https://huggingface.co/datasets/magicslabnu/GenoAdv}{Datasets} and can be accessed directly through their platform.

\section{Boarder Impact}
\label{ap:border}
This paper seeks to advance the trustworthiness of genomic foundation models (GFMs). 
While the work does not have immediate social implications, it represents a step toward creating more reliable GFMs. 
However, the adversarial samples released in the \textbf{GenoAdv} dataset and experiments can provide incorrect classification for existing GFMs.

\begin{figure}[htbp]
\centering
\hspace*{-15mm}
\resizebox{\textwidth}{!}{
\begin{tikzpicture}[
    every node/.style={draw, rounded corners=5pt, font=\sffamily},
    level 1/.style={sibling distance=55mm, level distance=18mm},
    level 2/.style={sibling distance=40mm, level distance=18mm},
    level 3/.style={sibling distance=25mm},
    edge from parent/.style={draw, -latex, to path={
        -- ++(0,-5pt) -| (\tikztotarget)}},
    main/.style={fill=white, minimum width=30mm, text width=30mm, align=center, font=\sffamily\bfseries},
    box/.style={fill=white, minimum width=25mm, text width=25mm, align=center},
    ref/.style={fill=blue!10, rounded corners=5pt, text width=55mm, align=left, font=\sffamily\small\color{blue}}
]
\node[main, font=\rmfamily, rotate=90, minimum height=15mm, text width=50mm] at (-3.5,-4) {Adversarial Strategies};
\node[main, font=\rmfamily, text width=18mm, align=center] (attack) at (0,0) {Attack\\Methods};
\node[main, font=\rmfamily, text width=18mm, align=center] (defense) at (0,-8) {Defense\\Methods};
\draw[line width=1.5pt] (-2.75,-4) -- (-2.25,-4);
\draw[line width=1.5pt] (-2.25,0) -- (-2.25,-8);
\draw[line width=1.5pt] (-2.25,0) -- (-1.5,0);
\draw[line width=1.5pt] (-2.25,-8) -- (-1.5,-8);
\node[main, font=\rmfamily] (untarg) at (6,3) {Untargeted};
\node[main, font=\rmfamily] (targ)   at (6,0) {Targeted};
\node[main, font=\rmfamily] (univ)   at (6,-3){Universal};
\node[main, font=\rmfamily] (defense_train)   at (6,-7)   {Adversarial\\Training};
\node[main, font=\rmfamily] (defense_dist)    at (6,-9.5) {Defensive\\Distillation};
\node[main, font=\rmfamily] (defense_detect)  at (6,-12)  {Sample\\Detection};
\node[main, font=\rmfamily] (defense_regular) at (6,-14.5){Regularization,\\Purification,\\Certified Robustness};
\draw[line width=1.5pt] (2.8,3) -- (2.8,-3);
\draw[line width=1.5pt] (1.5,0) -- (2.8,0);
\draw[line width=1.5pt] (2.8,3) -- (4.35,3);
\draw[line width=1.5pt] (2.8,0) -- (4.35,0);
\draw[line width=1.5pt] (2.8,-3) -- (4.35,-3);

\draw[line width=1.5pt] (2.8,-14.5) -- (2.8,-7);
\draw[line width=1.5pt] (1.5,-8) -- (2.8,-8);
\draw[line width=1.5pt] (2.8,-7) -- (4.35,-7);
\draw[line width=1.5pt] (2.8,-9.5) -- (4.35,-9.5);
\draw[line width=1.5pt] (2.8,-12) -- (4.35,-12);
\draw[line width=1.5pt] (2.8,-14.5) -- (4.35,-14.5);
\node[ref, font=\rmfamily, right=8mm of untarg] (ref1){\citet{yu2025boost,FGSM,wu2019untargeted,kurakin2018adversarial,madry2017towards,moosavi2015deepfool}};
\node[ref, font=\rmfamily, right=8mm of targ]   (ref2){\citet{zhang2024lp,li2020towards,di2020taamr,chen2018ead,wiyatno2018maximaljacobianbasedsaliencymap,carlini2016towards}};
\node[ref, font=\rmfamily, right=8mm of univ]   (ref3){\citet{skovorodnikov2024fimba,ye2023fg,zhang2021data, poursaeed2018generative,mopuri2018nagnetworkadversarygeneration,khrulkov2018geometryscoremethodcomparing,moosavi2017universal}};
\node[ref, font=\rmfamily, right=8mm of defense_train]   (ref4){\citet{adfar, Zhu2020FreeLB, zheng2020efficient, pgd}};
\node[ref, font=\rmfamily, right=8mm of defense_dist]    (ref5){\citet{elgamrani2024adversarial, papernot2016distillation}};
\node[ref, font=\rmfamily, right=8mm of defense_detect]  (ref6){\citet{cohen2020detecting,wang2019adversarial,lee2018simple,feinman2017detect}};
\node[ref, font=\rmfamily, right=8mm of defense_regular] (ref7){\citet{xuan2025exploringimpacttemperaturescaling,yang-etal-2023-label-smoothing,liu-etal-2022-flooding,li-etal-2023-text,jia-etal-2019-certified}};
\draw[line width=1.5pt] (untarg) -- (ref1);
\draw[line width=1.5pt] (targ) -- (ref2);
\draw[line width=1.5pt] (univ) -- (ref3);
\draw[line width=1.5pt] (defense_train) -- (ref4);
\draw[line width=1.5pt] (defense_dist) -- (ref5);
\draw[line width=1.5pt] (defense_detect) -- (ref6);
\draw[line width=1.5pt] (defense_regular) -- (ref7);
\end{tikzpicture}
}
\caption{\textbf{Taxonomy of Adversarial Strategies.}}
\label{fig:adversarial_tree_styled}
\end{figure}

\section{Related Work}
\label{ap:background}
In this section, we explore the background of vulnerabilities in GFMs. We begin by introducing benchmarks for evaluating adversarial attacks on GFMs, including standard datasets, metrics, and evaluation protocols. Next, we review existing adversarial attack methods tailored for GFMs, such as BERT-Attack \citep{li2020bert} and PGD \citep{pgd}. Finally, we discuss defense strategies against these attacks, covering approaches like FreeLB \citep{Zhu2020FreeLB} and ADFAR \citep{adfar}.

\subsection{Benchmarks}
The GUE benchmark \citep{zhou2024dnabert2} encompasses a variety of genome classification tasks, including promoter detection, transcription factor prediction, and COVID variant classification. These tasks are designed to assess model performance across multiple species, such as humans, fungi, viruses, and yeast. Building on this, GUE+ extends the benchmark to focus on tasks involving longer input sequences, ranging from 5000 to 10000 base pairs, to evaluate models’ capabilities in processing and analyzing complex genomic data. The GUE benchmark assesses model performance using metrics such as Accuracy, F1-score, and Matthews Correlation Coefficient (MCC) \citep{chicco2020advantages}. 

Meanwhile, GenBench \citep{liu2024genbench} is a comprehensive benchmarking suite tailored for evaluating the performance of GFMs. It systematically analyzes datasets from diverse biological domains, with a focus on both short-range and long-range genomic tasks. These tasks encompass essential areas such as coding regions, non-coding regions, and genome structure. For classification tasks, GenBench uses cross-entropy loss to measure prediction divergence and evaluates performance with top-1 accuracy and AUC-ROC. For regression tasks, it applies Mean Squared Error (MSE) for accuracy and calculates Spearman and Pearson correlation coefficients to assess relationships.

These benchmarks \citep{liu2024genbench,grevsova2023genomic} offer a thorough evaluation of GFMs.
However, all these benchmarks overlook the safety aspects of the GFMs. 
Recently, the safety of large scientific foundation models has become a prominent focus in research \citep{li2024scisafeeval,skovorodnikov2024fimba}. As a groundbreaking approach to incorporating adversarial attacks into genomic data analysis, FIMBA \citep{skovorodnikov2024fimba} leverages publicly available genomic datasets, such as The Cancer Genome Atlas (TCGA) and COVID-19 single-cell RNA sequencing data, to assess the robustness of AI models against adversarial feature importance attacks.
 In the TCGA dataset, the classification task aims to determine whether a sample is malignant, while in the COVID-19 dataset, the objective is to identify whether a patient is diagnosed with the disease. As part of this evaluation, FIMBA uses Accuracy as the primary performance metric to measure the classification capability. To assess the quality and stealth of the adversarial attacks, they employ the Structural Similarity Index Measure (SSIM). SSIM quantifies the structural similarity between the original and adversarially attacked data, with higher values indicating attacks that are more undetectable and preserve the data's original structure.

\subsection{Adversarial Attack}
Adversarial attacks can be broadly classified into untargeted, targeted, and universal attacks. Untargeted attacks \citep{yu2025boost,FGSM,wu2019untargeted,kurakin2018adversarial,madry2017towards,moosavi2015deepfool} aim to cause any misprediction by modifying the input in the direction of the loss gradient, maximizing overall loss. In contrast, targeted attacks \citep{zhang2024lp,li2020towards,di2020taamr,chen2018ead,wiyatno2018maximaljacobianbasedsaliencymap,carlini2016towards} guide the model's output toward a specific attacker-defined class using the loss gradient directed at the target class. Universal attacks \citep{skovorodnikov2024fimba,ye2023fg,zhang2021data, poursaeed2018generative,mopuri2018nagnetworkadversarygeneration,khrulkov2018geometryscoremethodcomparing,moosavi2017universal} generate perturbations applicable to any input from a given class, causing mispredictions universally.

The Fast Gradient Sign Method (FGSM) \citep{FGSM} and Projected Gradient Descent (PGD) \citep{pgd} are two prominent techniques for generating adversarial examples in machine learning, particularly for deep neural networks \citep{shayegani2023surveyvulnerabilitieslargelanguage}. FGSM generates adversarial samples by applying a single-step perturbation in the direction of the gradient of the loss function, scaled to a predefined magnitude, making it computationally efficient. However, PGD improves robustness by iteratively applying small gradient-based perturbations while ensuring that adversarial examples remain within a specified norm constraint, leading to more effective attacks.

A variety of adversarial attack and defense strategies have recently been proposed, specifically tailored for natural language processing (NLP) tasks \citep{goyal2023survey}. These techniques can be categorized into character-level, word-level, and sentence-level adversarial attacks. 
Character-level adversarial attacks involve perturbing individual characters in text to mislead machine learning models while preserving readability. For example, DeepWordBug \citep{gao2018black} modifies specific characters based on importance scores to maximize the model's misclassification while minimizing changes to the text. Similarly, TextBugger \citep{li2018textbugger} generates adversarial examples by replacing, inserting, or removing characters, focusing on semantic preservation and evading detection by defense mechanisms.
Word-level adversarial attacks focus on perturbing entire words rather than individual characters. These attacks can be broadly classified into three categories: gradient-based, importance-based, and replacement-based methods. Gradient-based methods, such as FGSM \citep{FGSM}, utilize gradients to identify vulnerable words and modify them to maximize the model's loss. Importance-based methods, exemplified by TextFooler \citep{jin2020bert}, rank words based on their contribution to the model's prediction and replace them with semantically similar alternatives to alter the output. Replacement-based methods, like BERT-Attack \citep{li2020bert}, leverage pre-trained language models to generate context-aware substitutions, ensuring the adversarial examples maintain fluency and semantic coherence.
Sentence-level adversarial attacks involve generating adversarial examples by modifying entire sentences to mislead the model while maintaining grammaticality and semantic relevance. AdvGen \citep{cheng2019robust} generates adversarial sentences by leveraging reinforcement learning to iteratively modify sentence structures and word choices, ensuring the adversarial examples remain coherent and natural while effectively deceiving the target model.

Adversarial attacks have also been explored in genomic models to assess their robustness and identify vulnerabilities in sequence-based predictions. FIMBA \citep{skovorodnikov2024fimba} presents a black-box, model-agnostic attack and analysis framework designed for widely used machine learning models in genomics.
FIMBA targets genomic models by perturbing key features identified through SHAP values, which measure the importance of each feature to the model's decision. By selecting the most impactful features and modifying them using interpolation between the original and target vectors, FIMBA generates minimally altered adversarial examples that effectively deceive the model. The attack avoids gradient reliance, functioning as a black-box method, and focuses on modifying as few features as possible to ensure both high efficacy and low detectability.

\subsection{Attack Method Categorization}

To provide a comprehensive evaluation, we categorize our adversarial attack methods as follows:
\begin{itemize}
    \item \textbf{White-box attacks:}
    \begin{itemize}
        \item PGD
        \item BertAttack and TextFooler
        \item AutoAttack (ensemble of gradient-based methods)
        \item UAP
    \end{itemize}
    \item \textbf{Black-box attacks:}
    \begin{itemize}
        \item FIMBA
    \end{itemize}
    \item \textbf{Targeted vs. Untargeted settings:}\\
    All attacks except PGD and AutoAttack are performed in the untargeted setting. PGD and AutoAttack are evaluated in both targeted and untargeted modes.
\end{itemize}

\subsection{Defense Methods}
To improve the robustness of GFMs, various defense strategies \citep{ke2025detection, luo2024decoupled, adfar, Zhu2020FreeLB, cohen2020detecting,lee2018simple,papernot2016distillation} are proposed, including adversarial training, defensive distillation, adversarial sample detection, and regularization, purification, and certified robustness. Among these, adversarial training \citep{adfar, Zhu2020FreeLB, zheng2020efficient, pgd} is the most effective, enhancing model resilience by injecting adversarial examples during training. Among these methods, \citet{madry2017towards} propose a method to inject bounded perturbations into word embeddings and minimize worst‑case loss, almost halving BERT‑Attack and TextFooler success rates without degrading clean accuracy. FreeLB \citep{Zhu2020FreeLB} merges several PGD steps into one forward‑backward pass and accumulates gradients, cutting training cost; FreeLB++ \citep{li2021freelbpp} enlarges the radius and steps for further robustness gains at no extra accuracy loss. Other lightweight variants such as SMART\citep{jiang2020smart}, TAVAT \citep{li2021tavat}, and R3F \citep{aghajanyan2020better} approximate the inner maximization with uncertainty‑ or noise‑based regularization, reaching performance close to FreeLB++ at a fraction of the compute. The frequency‑aware randomization framework ADFAR \citep{adfar} incorporates anomaly‑detection signals and word‑frequency constraints directly into the training loop, unifying adversarial sample detection ideas with adversarial training to further weaken substitution‑based attacks without extra overhead. 
Defensive distillation \citep{elgamrani2024adversarial, papernot2016distillation} trains a student model on softened outputs from a teacher model to smooth decision boundaries, though its efficacy against strong adversarial attacks remains debated. However, \citet{carlini2016defensive} demonstrate that defensive distillation is ineffective against adaptive adversarial attacks, as carefully crafted inputs can still bypass the smoothed decision boundaries and fool the model. 
Adversarial sample detection \citep{cohen2020detecting,wang2019adversarial,lee2018simple,feinman2017detect} focuses on identifying malicious inputs rather than improving model robustness. MAFD \citep{Jin2024MAFD} combines perplexity, word frequency, and masking-probability features for robust anomaly scoring; ONION \citep{qi2021onion} leverages language-model perplexity to prune high-risk tokens; Sharpness-based detectors \citep{zheng2023detecting} add infinitesimal noise and flag samples exhibiting steep loss increases. Deployed alongside adversarial training, these detectors offer real‑time protection against unseen or cross‑domain attacks.
Regularization, purification and certified Robustness reduce perturbation sensitivity by modifying the loss or sanitizing inputs. Flooding‑X \citep{liu-etal-2022-flooding} maintains a loss floor to guide the model toward flatter regions; adversarial label smoothing \citep{yang-etal-2023-label-smoothing} and temperature scaling \citep{xuan2025exploringimpacttemperaturescaling} curb over‑confidence; masked‑language‑model purification \citep{li-etal-2023-text} masks and reconstructs suspicious tokens to cleanse perturbations. Interval bound propagation (IBP) \citep{jia-etal-2019-certified} and randomized smoothing schemes such as SAFER \citep{ye-etal-2020-safer} and RanMASK \citep{zeng2021ranmask} provide formal guarantees against word substitutions or masking budgets.

\section{Ethical Considerations}
\label{ap:ethical}
Prior to making this work public, we share our adversarial attack artefacts
and our results with leading GFMs teams, as shown in \cref{ap:disclosure}. Secondly, we open-source the code and data used in our experiments to promote transparency.
Also, we carefully consider the ethical impact of our work and list the two impacts: (1) The adversarial sample released in the \textbf{GenoAdv} dataset and experiments can provide incorrect classification for existing GFMs. (2) Adversarial training is an efficiency method to make GFMs more resilient to adversarial attacks.

\subsection{Dual-Use and Misuse Risks}
\label{ap:dualuse}

We recognize that adversarial attacks on genomic foundation models (GFMs), particularly those applied to clinical diagnostics and gene pathogenicity prediction, raise significant dual-use and misuse concerns. While our intention is to improve the safety and robustness of GFMs, we acknowledge that, if misused, the techniques developed in this work could be repurposed to evade genomic screening, manipulate diagnostic predictions, or interfere with treatment decision-making.

The adversarial samples included in the \textbf{GenoAdv} dataset are designed to reveal vulnerabilities in current models by targeting biologically meaningful regions. These vulnerabilities highlight the urgency for robust defensive strategies. However, we also recognize that releasing such resources without caution could present opportunities for malicious use.

To mitigate these risks, we take the following steps. First, we have contacted several leading GFM development teams to disclose our findings and foster collaboration on model hardening. Second, although we open-sourced our code and data to promote reproducibility, we now include a usage statement specifying that the tools and dataset are intended strictly for non-commercial research purposes. All data directly generated from adversarial attacks will not be publicly released and will be securely stored; any organizations or individuals seeking access must directly contact the authors, and requests will be evaluated by healthcare cybersecurity professionals within our team. Use in clinical or diagnostic applications, or for purposes that could impact public health, is explicitly discouraged.

We urge future researchers to approach this line of work with similar responsibility. Any use of GenoAdv or our attack pipeline should be guided by ethical principles that prioritize model reliability, biosecurity, and societal benefit. Our overarching goal is not to facilitate harm, but to proactively identify and close security gaps in genomic models before they can be exploited in real-world settings.

\section{Reproducibility}
\label{ap:reproduciblity}
In this section, we provide a discussion on the reproducibility of our experiments, including the details of the datasets used, the training and evaluation protocols, and the hyperparameters employed in our experiments.

\subsection{Source of Randomness.} To ensure reproducibility, we run all experiments using three different random seeds. We observe that the results are highly stable, with the benchmark introducing only minor variations—showing a variance of at most 2\%.

\subsection{Implementation.} 
To ensure reproducibility, we implement the adversarial attack and defense methods based on their official GitHub repositories, as shown below:
\begin{itemize}
    \item \textbf{BertAttack:} https://github.com/LinyangLee/BERT-Attack
    \item \textbf{TextFooler:} https://github.com/jind11/TextFooler
    \item \textbf{PGD:} https://github.com/MadryLab/robustness
    \item \textbf{FIMBA:} https://github.com/HeorhiiS/fimba-attack
    \item \textbf{ADFAR:} https://github.com/LilyNLP/ADFAR
    \item \textbf{FreeLB:} https://github.com/zhuchen03/FreeLB
    \item \textbf{AutoAttack:} https://github.com/fra31/auto-attack
    \item \textbf{Universal:} https://github.com/LTS4/universal

\end{itemize}

\subsection{Hyperparameter.} 
\label{ap:hyperparameter}
We present the hyperparameters used in the benchmark for each model. 
We use \textbf{AdamW} \citep{loshchilov2017decoupled} as the optimizer. 
Fine-tuning and adversarial training are performed uniformly across all models and datasets for 4 epochs, using a batch size of 64 and a maximum sequence length of 256. We use the AdamW optimizer with a learning rate of $3e^{-5}$, gradient accumulation steps of 1, and a warmup ratio of 0.05.
The maximum sequence length and batch size used for each adversarial attack and defense method are summarized in Table~\ref{tab:hyperparams_horizontal}. These settings are chosen to balance computational efficiency and attack effectiveness across different methods.

\begin{table}[h]
\centering
\small
\caption{Hyperparameter settings for each attack method.}
\label{tab:hyperparams_horizontal}
\begin{tabular}{lcccccc}
\toprule
\textbf{Hyperparameter} 
& \textbf{BertAttack} 
& \textbf{TextFooler} 
& \textbf{PGD} 
& \textbf{FIMBA} 
& \textbf{ADFAR} 
& \textbf{FreeLB} \\
\midrule
Max Sequence Length 
& 128 
& 256 
& 256 
& 128 
& 128 
& 256 \\
Batch Size 
& 32 
& 128 
& 16 
& 32 
& 2 
& 32 \\
\bottomrule
\end{tabular}
\end{table}

For \textbf{BertAttack}, we configure the attack with \texttt{k} = 48 and set the prediction score threshold to 0, using DNABERT-2 as the reference masked language model.
In \textbf{ADFAR’s} frequency-aware randomization process, we set the frequency threshold $f_{\text{thres}} = 200$, the number of samples $n_s = 20$, and the number of features $n_f = 10$.
For \textbf{FreeLB}, the hyperparameters used in our experiments include an adversarial learning rate of 0.1, adversarial magnitude of 0.6, two adversarial steps, a base learning rate of $1e^{-5}$, gradient accumulation steps set to 1, and a weight decay of $1e^{-2}$.

\section{Additional \sys demonstration}
\label{ap:demo}
We provide two installation options for \sys and two usage methods: via command line and Python code.

\begin{tcolorbox}[colback=red!5, colframe=black!60, boxrule=0.5pt, arc=2pt, title=Example of Installation of \sys ,floatplacement=h]
\begin{verbatim}
# Install with pip
pip install genoarmory

# Install with source code
git clone https://github.com/MAGICS-LAB/GenoArmory.git
conda create -n genoarmory pip=3.9
pip install .

\end{verbatim}
\end{tcolorbox}

\begin{tcolorbox}[colback=orange!5, colframe=black!60, boxrule=0.5pt, arc=2pt, title=Example of Python Usage of \sys ,floatplacement=h]
\begin{verbatim}
# Initialize model
from GenoArmory import GenoArmory
import json
# You need to initialize GenoArmory with a model and tokenizer.
gen = GenoArmory(model=None, tokenizer=None)
params_file = 'xxx/scripts/PGD/pgd_dnabert.json'

# Visulization
gen.visualization(
    folder_path='xxx/BERT-Attack/results/meta/test',
    output_pdf_path='xxx/BERT-Attack/results/meta/test'
)

# Attack
if params_file:
  try:
      with open(params_file, "r") as f:
          kwargs = json.load(f)
  except json.JSONDecodeError as e:
      raise ValueError(f"Invalid JSON in params file")
  except FileNotFoundError:
      raise FileNotFoundError(f"Params file not found.")

gen.attack(
    attack_method='pgd',
    model_path='magicslabnu/GERM',
    **kwargs
)
\end{verbatim}
\end{tcolorbox}

\begin{tcolorbox}[colback=blue!5, colframe=black!60, boxrule=0.5pt, arc=2pt, title=Example of Commend Line Usage of \sys ,floatplacement=h]
\begin{verbatim}
# Attack
python GenoArmory.py 
--model_path magicslabnu/GERM attack 
--method pgd --params_file xxx/scripts/PGD/pgd_dnabert.json


# Defense
python GenoArmory.py 
--model_path magicslabnu/GERM defense 
--method at --params_file xxx/scripts/AT/at_pgd_dnabert.json

# Visualization
python GenoArmory.py 
--model_path magicslabnu/GERM visualize 
--folder_path xxx/BERT-Attack/results/meta/test 
--save_path xxx/BERT-Attack/results/meta/test/frequency.pdf


# Read MetaData
python GenoArmory.py 
--model_path magicslabnu/GERM read 
--type attack --method TextFooler --model_name dnabert

\end{verbatim}
\end{tcolorbox}

\section{Disclosure}
\label{ap:disclosure}
We share our disclosure with the authors of DNABERT-2, NT, HyenaDNA, and GenomeOcean to inform them of our findings and benchmark. Also, we highlight the potential impact on their models in our disclosure. 

\begin{tcolorbox}[colback=black!5!white,colframe=black,title=Example of Disclosure Letter\label{box:dislosure},floatplacement=h]
Dear DNABERT/DNABERT-2/DNABERT-S team,

\texttt{
We hope this message finds you well. We are reaching out to share the preliminary results and artifacts from our recent study on adversarial attacks targeting DNA-based Genomic Foundation Models (GFMs), which we plan to release publicly as part of a unified benchmarking framework.}

\texttt{Given your leading role in the development of GFMs, we believe it is essential to disclose our findings to you in advance. Our results demonstrate that carefully crafted adversarial sequences can induce incorrect classifications across multiple GFM architectures. We also find that adversarial training remains a promising defense strategy for enhancing model robustness.}

\texttt{To support responsible disclosure, we are providing:}

\texttt{1. A summary of key findings and model vulnerabilities}

\texttt{2. The adversarial sample set and evaluation scripts}

\texttt{3. A description of our ethical considerations and intended safeguards}

\texttt{We welcome your feedback on potential risks, mitigation strategies, and collaborative opportunities to ensure this research contributes constructively to the GFM community.}

\texttt{Please let us know if you would like early access to the materials or would prefer to schedule a meeting to discuss further.}

Best regards,

\sys Author

\end{tcolorbox}

\section{Disclosure of LLM Usage}
\label{ap:llm}
We utilize Cursor to assist in writing repetitive bash automation scripts and employ GPT-4o to refine the paper's language for conciseness and precision.

\section{Experiment Setting}
\subsection{Computational Resource}
\label{ap:resource}
We perform all experiments using 4 NVIDIA H100 GPUs with 80GB of memory and a 24-core Intel(R) Xeon(R) Gold 6338 CPU operating at 2.00 GHz.

\subsection{Metrics of Experiments}
\label{ap:metrics}

In our experiments, we use two core metrics to evaluate the effectiveness of adversarial attacks and the robustness of defense strategies: \textbf{Attack Success Rate (ASR)} and \textbf{Defense Success Rate (DSR)}.

\textbf{Attack Success Rate (ASR)} is defined as the relative drop in accuracy caused by the adversarial attack. Formally, let $A_{\text{clean}}$ be the model accuracy on clean inputs and $A_{\text{adv}}$ be the accuracy on adversarial inputs, then:
\begin{align}
    \text{ASR} = \frac{A_{\text{clean}} - A_{\text{adv}}}{A_{\text{clean}}} \times 100\%.
\end{align}

\textbf{Defense Success Rate (DSR)} measures the robustness gain achieved by applying a defense mechanism. Let $A_{\text{def}}$ be the accuracy of the defended model on adversarial inputs, then:
\begin{align}
    \text{DSR} = (1- \frac{A_{\text{def}} - A_{\text{adv}}}{A_{\text{def}}}) \times 100\%.
\end{align}

These metrics allow us to quantitatively assess both the impact of adversarial attacks and the degree to which defenses can mitigate that impact.

\subsection{Implementation}
\label{ap:implementation}
For DNABERT-2, we use the 117-million-parameter version of the model\footnote{\href{https://huggingface.co/zhihan1996/DNABERT-2-117M}{zhihan1996/DNABERT-2-117M}}. For NT, we use the 2.5-billion-parameter version of the model\footnote{\href{https://huggingface.co/InstaDeepAI/nucleotide-transformer-2.5b-multi-species}{InstaDeepAI/nucleotide-transformer-2.5b-multi-species}}. For NT2, we use the 100-million-parameter version of the model\footnote{\href{https://huggingface.co/InstaDeepAI/nucleotide-transformer-v2-100m-multi-species}{InstaDeepAI/nucleotide-transformer-v2-100m-multi-species}}.  
For HyenaDNA, we use the 4.07-million-parameter version of the model\footnote{\href{https://huggingface.co/LongSafari/hyenadna-small-32k-seqlen-hf}{LongSafari/hyenadna-small-32k-seqlen-hf}}. 
All four models represent state-of-the-art approaches for genome sequence classification tasks, consistently achieving high performance across various datasets.
GenomeOcean \citep{zhou2025genomeocean}, on the other hand, is a transformer-based model designed explicitly for genome sequence generation tasks, demonstrating superior performance compared to existing models, such as Evo \citep{nguyen2024sequence}.
We use the 100-million-parameter version of the model\footnote{\href{https://huggingface.co/pGenomeOcean/GenomeOcean-100M}{pGenomeOcean/GenomeOcean-100M}}.
For our experiments, we fine-tuned all of these models using their official checkpoints on the datasets employed in this study. 

\subsection{Downstream Tasks Across Different Models}
\label{ap:downstream}
We examine the downstream tasks of several genomic foundation models (GFMs), including DNABERT-2 \citep{zhou2024dnabert2}, HyenaDNA \citep{nguyen2024hyenadna}, GenomeOcean \citep{zhou2025genomeocean}, and Nucleotide Transformer \citep{dalla2024nucleotide}. As summarized in \cref{tab:models_tasks}, these models primarily focus on classification tasks. In contrast, our analysis of the GenBench datasets \citep{liu2024genbench} reveals the inclusion of regression tasks, offering a more comprehensive evaluation framework.

\begin{table}[h!]
    \centering
     \caption{\textbf{Comparison of Models (Benchmarks) and Their Tasks.}}
    \label{tab:models_tasks}
    \resizebox{\textwidth}{!}{
    \begin{tabular}{@{}llc@{}}
        \toprule
        Model    & Tasks  & Classification-Only \\ \midrule
        DNABERT-2            & GUE (28 Classification tasks)                                         & Yes                 \\
        Nucleotide Transformer & Nucleotide Transformer Benchmark (18 Classification tasks)           & Yes                 \\
        HyenaDNA             & GenBench (Classification-Only) + Nucleotide Transformer Benchmark     & Yes                 \\
        GenomeOcean & Classification + Generation (5 GUE Classification tasks) & No \\
        GenBench             & Classification + Regression (e.g., Drosophila Enhancer Activity Prediction) & No                  \\ 
        \bottomrule
    \end{tabular}
   }
\end{table}

\section{Additional Numerical Experiments}

\subsection{Additional Experiments on Universal and AutoAttack}

In this section, we present additional experimental results to further evaluate the robustness of the target models and defense strategies under adversarial attacks. Specifically, we introduce two representative attack method AutoAttack and Universal Attack, to comprehensively assess model performance. The experiments include attack success rates (ASR), defense success rates (DSR) under various defense techniques, and evaluation of model augmentation using the GenoAdv dataset. These results provide deeper insights into the effectiveness of different strategies against adversarial perturbations and demonstrate the impact of data augmentation on adversarial robustness.

\begin{table}[ht]
  \centering
  \caption{\textbf{Additional Adversarial Attack Performance of the Evaluated Method.} 
  We conduct additional experiments to assess the effectiveness of the Auto Attack and Universal Attack against target models. The table presents a comparison of target model performance before and after applying the evaluated attack. We report Attack Success Rate (ASR) as the primary evaluation metric, with variance omitted as they are all $\le 2$\%.}
  \label{tab:addational_attack}
  \resizebox{0.9\textwidth}{!}{
    \begin{tabular}{lccccc}
      \toprule
      &  \multicolumn{4}{c}{Transformer-based} & Hyena-based \\
      \cmidrule(r){2-5} \cmidrule(l){6-6}
      Attack  & DNABERT-2 & NT & NT2 & GenomeOcean & HyenaDNA \\
      \midrule
      Universal & 53.19 & 53.19 & 53.19 & 53.19 & 53.19 \\
      Auto      & 50.27 & 50.27 & 50.27 & 50.27 & 50.27 \\
      \bottomrule
    \end{tabular}
  }
\end{table}

\begin{table}[ht]
  \centering
  \caption{\textbf{Additional Defense Performance Under Adversarial Attacks.} We conducted additional experiments to evaluate the performance of a defense method against adversarial attacks. 
  The table compares the performance of target models, both with and without the evaluated defense, under Auto and Universal Attack. The Defense Success Rate (DSR) is used as the primary evaluation metric, with variance omitted as they are all $\le 2$\%. In the table, \textbf{AT} denotes traditional adversarial training.}
  \label{tab:addational_defense}
  \resizebox{0.9\textwidth}{!}{
  \begin{tabular}{lcccccc}
    \toprule
    & & \multicolumn{4}{c}{Transformer-based} & Hyena-based \\
    \cmidrule(r){3-6} \cmidrule(l){7-7}
    Attack Method & Defense  & DNABERT-2 & NT & NT2 & GenomeOcean & HyenaDNA\\
    \midrule
    \multirow{4}{*}{Universal} 
       & N/A    & 53.19 & 53.68 & 55.43 & 54.53 & 67.83 \\
       & AT     & 54.33 & 53.53 & 55.73 & 55.55 & 67.62 \\
       & FreeLB & 56.56 & 55.44 & 56.67 & 55.96 & 68.07 \\
       & ADFAR  & 55.77 & 54.58 & 56.21 & 54.41 & 68.17 \\
    \midrule
    \multirow{4}{*}{Auto}   
       & N/A    & 50.27 & 51.26 & 54.09 & 47.83 & 41.58 \\
       & AT     & 52.13 & 51.88 & 53.27 & 49.64 & 42.28 \\
       & FreeLB & 51.97 & 52.21 & 54.16 & 48.02 & 42.72 \\
       & ADFAR  & 52.20 & 52.39 & 54.26 & 47.88 & 42.51 \\
    \bottomrule
  \end{tabular}
  }
\end{table}

\begin{table}[ht]
\centering
\caption{\textbf{Additional Defense Performance Augmented with the GenoAdv Dataset.}
We conduct additional experiments to evaluate the performance of a model augmented with the GenoAdv dataset against adversarial attacks.
The table compares the performance of the target models, both with and without the GenoAdv dataset augmentation, under several attack methods and defense strategies.
We report accuracy as the primary evaluation metric. The best results are highlighted in bold. In the table, \textbf{AT} denotes traditional adversarial training.
}
\label{tab:addational_augmented}
\resizebox{0.9\textwidth}{!}{
  \begin{tabular}{lcccccc}
    \toprule
    & & \multicolumn{4}{c}{Transformer-based} & Hyena-based \\
    \cmidrule(r){3-6} \cmidrule(l){7-7}
    Attack Method & Defense  & DNABERT-2 & NT & NT2 & GenomeOcean & HyenaDNA\\
    \midrule
    \multirow{3}{*}{Universal}
    &N/A      & 53.19 & 53.68 & 55.43 & 54.53 & 67.83 \\
    &AT        & 54.33 & 53.53 & 55.73 & 55.55 & 67.62 \\
    &GenoAdv   & \textbf{57.23} & \textbf{55.11} & \textbf{56.36} & \textbf{55.54} & \textbf{70.06} \\
    \midrule
    \multirow{3}{*}{Auto}
    &N/A      & 50.27 & 51.26 & 54.09 & 47.83 & 41.58 \\
    &AT        & 52.13 & 51.88 & 53.27 & 49.64 & 42.28 \\
    &GenoAdv   & \textbf{54.44} & \textbf{53.35} & \textbf{54.79} & \textbf{52.41} & \textbf{48.23} \\
    \bottomrule
  \end{tabular}
}
\end{table}

\subsection{All results in Adversarial Attack}
This section provides a comprehensive evaluation of multiple adversarial attacks across different GFM models. We compare BertAttack, TextFooler, FIMBA, and PGD on a range of biolGenomeOceanical prediction tasks, including epigenetic marks prediction, promoter detection, and transcription factor prediction in both human and mouse datasets. The evaluated GFM models include DNABERT-2, NT, NT2, HyenaDNA, and GenomeOcean.

\begin{table}[ht]
    \centering
    \caption{\textbf{Performance Comparison of Adversarial Attacks on DNABERT-2.} This table shows the performance of all adversarial attacks on the DNABNERT-2 model. All results are evaluated using the Attack Success Rate (ASR) metric. The best result is highlighted in bold, while the second-best
result is underlined. }
    \begin{minipage}{\textwidth}
    \centering
    \resizebox{0.7\textwidth}{!}{%
\begin{tabular}{lccccccc}
        \toprule
        & \multicolumn{6}{c}{Epigenetic Marks Prediction} \\
        \cmidrule(lr){2-7}
        Attack & H3 & H3K14ac & H3K36me3 & H3K4me1 & H3K4me2 & H3K4me3 \\
        \midrule
        BertAttack & \textbf{91.20} & \underline{99.70} & \underline{99.80} & \textbf{95.10} & \textbf{99.20} & \underline{99.30} \\
        TextFooler & \underline{90.40} & \textbf{99.90} & \textbf{99.90} & \underline{86.50} & \textbf{99.20} & \textbf{100.00} \\
        FIMBA & 43.70 & 51.90 & 24.00 & 41.30 & 26.90 & 41.70 \\
        PGD & 41.30 & 33.30 & 35.50 & 35.90 & 38.40 & 31.80 \\
        \bottomrule
    \end{tabular}
    }
    \end{minipage}
   \begin{minipage}{\textwidth}
    \centering
    \resizebox{0.7\textwidth}{!}{%
    \begin{tabular}{lccccccc}
        \toprule
        & \multicolumn{4}{c}{Epigenetic Marks Prediction} & \multicolumn{3}{c}{Promoter Detection (300bp)} \\
        \cmidrule(lr){2-5} \cmidrule(lr){6-8}
        Attack & H3K79me3 & H3K9ac & H4 & H4ac & all & notata & tata \\ %
        \midrule
        BertAttack & \underline{97.50} & \textbf{98.00} & \textbf{96.60} & \textbf{100.00} & \textbf{83.70} & \textbf{92.70} & \underline{96.50} \\
        TextFooler & \textbf{99.40} & \underline{96.20} & \underline{96.00} & \underline{94.20} & \underline{71.80} & 28.30 & \textbf{97.00} \\
        FIMBA & 24.40 & 43.80 & 36.60 & 50.60 & 58.30 & 14.90 & 87.10 \\
        PGD & 41.40 & 39.30 & 36.20 & 46.10 & 45.60 & \underline{43.50} & 42.90 \\
        \bottomrule
    \end{tabular}
    }
    \end{minipage}
    \begin{minipage}{\textwidth}
        \centering
    \resizebox{0.7\textwidth}{!}{%
    \begin{tabular}{lcccccccccc}
        \toprule
        &\multicolumn{5}{c}{Transcription Factor Prediction (Human)} & \multicolumn{3}{c}{Core Promoter Detection} \\
        \cmidrule(lr){2-6} \cmidrule(lr){7-9}
        Attack & tf0 & tf1 & tf2 & tf3 & tf4 & all & notata & tata \\ %
        \midrule
        BertAttack & \underline{96.80} & \underline{97.60} & \underline{99.80} & \underline{90.20} & \underline{97.40} & \textbf{99.20} & \textbf{99.30} & \textbf{98.90} \\
        TextFooler & 96.40 & \textbf{98.00} & \textbf{99.40} & \textbf{91.30} & \textbf{98.80} & \underline{97.40} & \underline{97.10} & \underline{92.00} \\
        FIMBA & 50.00 & 34.10 & 55.60 & 25.40 & 45.30 & 44.00 & 32.10 & 28.20 \\
        PGD & 36.60 & 32.30 & 35.60 & 34.80 & 41.00 & 35.10 & 34.10 & 35.80 \\
        \bottomrule
    \end{tabular}
    }
    \end{minipage} %
    \begin{minipage}{\textwidth}
        \centering
    \resizebox{0.55\textwidth}{!}{%
    \begin{tabular}{lcccccccc}
        \toprule
        &\multicolumn{5}{c}{Transcription Factor Prediction (Mouse)} \\
        \cmidrule(lr){2-6}
        Attack & 0 & 1 & 2 & 3 & 4  \\
        \midrule
        BertAttack & \underline{93.40} & \textbf{96.40} & \underline{96.20} & \underline{90.90} & \textbf{96.90}  \\
        TextFooler & \textbf{94.20} & \underline{94.50} & \textbf{97.20} & \textbf{92.40} & \underline{94.20} &  \\
        FIMBA & 46.40 & 3.10  & 43.30 & 46.40 & 39.50  \\
        PGD & 43.50 & 38.80 & 35.10 & 45.40 & 36.00 \\
        \bottomrule
    \end{tabular}
    }
    \end{minipage} %
\end{table}

\begin{table}[ht]
    \centering
    \caption{\textbf{Performance Comparison of Adversarial Attacks on HyenaDNA.} This table shows the performance of all adversarial attacks on the HyenaDNA model. All results are evaluated using the Attack Success Rate (ASR) metric. The best result is highlighted in bold, while the second-best result is underlined.}
    
    \begin{minipage}{\textwidth}
    \centering
    \resizebox{0.7\textwidth}{!}{%
    \begin{tabular}{lcccccc}
        \toprule
        & \multicolumn{6}{c}{Epigenetic Marks Prediction} \\
        \cmidrule(lr){2-7}
        Attack & H3 & H3K14ac & H3K36me3 & H3K4me1 & H3K4me2 & H3K4me3 \\
        \midrule
        BertAttack & \textbf{100.00} & \textbf{100.00} & \textbf{100.00} & \textbf{99.06} & \textbf{100.00} & \textbf{100.00} \\
        TextFooler & \textbf{100.00} & \textbf{100.00} & \textbf{100.00} & \underline{92.70} & \textbf{100.00} & \underline{91.14} \\
        FIMBA & 46.27 & 3.17 & 3.51 & 16.13 & 14.81 & 8.20 \\
        PGD & 10.70 & 6.70 & 91.14 & 5.11 & 90.68 & 4.45 \\
        \bottomrule
    \end{tabular}
    }
    \end{minipage}
    
    \begin{minipage}{\textwidth}
    \centering
    \resizebox{0.7\textwidth}{!}{%
    \begin{tabular}{lccccccc}
        \toprule
        & \multicolumn{4}{c}{Epigenetic Marks Prediction} & \multicolumn{3}{c}{Promoter Detection (300bp)} \\
        \cmidrule(lr){2-5} \cmidrule(lr){6-8}
        Attack & H3K79me3 & H3K9ac & H4 & H4ac & all & notata & tata \\
        \midrule
        BertAttack & \textbf{100.00} & \textbf{100.00} & \textbf{100.00} & \textbf{100.00} & \textbf{100.00} & \underline{97.06} & \textbf{100.00} \\
        TextFooler & \underline{35.79} & \underline{41.68} & \textbf{100.00} & \underline{99.19} & 46.49 & \textbf{99.19} & 92.85 \\
        FIMBA & 25.86 & 38.10 & 18.18 & 35.48 & \underline{48.68} & 31.17 & 41.67 \\
        PGD & 7.04 & 12.23 & 22.12 & 2.58 & 25.13 & 92.41 & \underline{93.72} \\
        \bottomrule
    \end{tabular}
    }
    \end{minipage}
    
    \begin{minipage}{\textwidth}
    \centering
    \resizebox{0.7\textwidth}{!}{%
    \begin{tabular}{lccccccccc}
        \toprule
        & \multicolumn{5}{c}{Transcription Factor Prediction (Human)} & \multicolumn{3}{c}{Core Promoter Detection} \\
        \cmidrule(lr){2-6} \cmidrule(lr){7-9}
        Attack & tf0 & tf1 & tf2 & tf3 & tf4 & all & notata & tata \\
        \midrule
        BertAttack & \textbf{100.00} & \underline{99.88} & \textbf{100.00} & \textbf{100.00} & \underline{98.81} & \textbf{100.00} & \textbf{100.00} & \textbf{100.00} \\
        TextFooler & \textbf{100.00} & \textbf{100.00} & \textbf{100.00} & \textbf{100.00} & \textbf{100.00} & \textbf{100.00} & \textbf{100.00} & \textbf{100.00} \\
        FIMBA & 38.16 & 35.71 & 31.94 & 26.39 & 48.86 & 34.15 & 32.14 & 33.33 \\
        PGD & 90.42 & 92.86 & 93.24 & 90.70 & 96.65 & 24.47 & 12.25 & 93.59 \\
        \bottomrule
    \end{tabular}
    }
    \end{minipage}
    
    \begin{minipage}{\textwidth}
    \centering
    \resizebox{0.55\textwidth}{!}{%
    \begin{tabular}{lccccc}
        \toprule
        & \multicolumn{5}{c}{Transcription Factor Prediction (Mouse)} \\
        \cmidrule(lr){2-6}
        Attack & 0 & 1 & 2 & 3 & 4 \\
        \midrule
        BertAttack & \textbf{100.00} & \underline{99.97} & \textbf{100.00} & \textbf{100.00} & \underline{98.79} \\
        TextFooler & 0.74 & \textbf{100.00} & \textbf{100.00} & \textbf{100.00} & \textbf{100.00} \\
        FIMBA & \underline{40.79} & 40.22 & 36.59 & 32.84 & 26.67 \\
        PGD & 0.00 & 4.35 & 2.65 & 90.99 & 90.18 \\
        \bottomrule
    \end{tabular}
    }
    \end{minipage}
\end{table}

\begin{table}[ht]
    \centering
    \caption{\textbf{Performance Comparison of Adversarial Attacks on NT.} This table shows the performance of all adversarial attacks on the Nucleotide Transformer (NT) model. All results are evaluated using the Attack Success Rate (ASR) metric. The best result is highlighted in bold, while the second-best
result is underlined.}
    \begin{minipage}{\textwidth}
    \centering
    \resizebox{0.7\textwidth}{!}{%
    \begin{tabular}{lccccccc}
        \toprule
        & \multicolumn{6}{c}{Epigenetic Marks Prediction} \\
        \cmidrule(lr){2-7}
        Attack & H3 & H3K14ac & H3K36me3 & H3K4me1 & H3K4me2 & H3K4me3 \\
        \midrule
        BertAttack & \textbf{99.92} & \textbf{100.00} & \textbf{100.00} & \textbf{100.00} & \textbf{100.00} & \textbf{100.00} \\
        TextFooler & \underline{66.23} & \textbf{100.00} & \underline{92.29} & \underline{97.32} & \textbf{100.00} & \textbf{100.00} \\
        FIMBA & 55.13 & 42.65 & 25.00 & 22.06 & 39.06 & 31.67 \\
        PGD & 38.53 & 38.45 & 39.11 & 36.16 & 36.93 & 25.25 \\
        \bottomrule
    \end{tabular}
    }
    \end{minipage}
   \begin{minipage}{\textwidth}
    \centering
    \resizebox{0.7\textwidth}{!}{%
    \begin{tabular}{lccccccc}
        \toprule
        & \multicolumn{4}{c}{Epigenetic Marks Prediction} & \multicolumn{3}{c}{Promoter Detection (300bp)} \\
        \cmidrule(lr){2-5} \cmidrule(lr){6-8}
        Attack & H3K79me3 & H3K9ac & H4 & H4ac & all & notata & tata \\ %
        \midrule
        BertAttack & \textbf{100.00} & \textbf{100.00} & \textbf{99.24} & \textbf{100.00} & \textbf{100.00} & \textbf{100.00} & \textbf{100.00} \\
        TextFooler & \textbf{100.00} & \textbf{100.00} & \underline{90.70} & \underline{89.24} & \underline{99.19} & \textbf{100.00} & \underline{91.20} \\
        FIMBA & 30.77 & 36.36 & 58.89 & 32.20 & 57.45 & 44.90 & 46.51 \\
        PGD & 40.91 & 20.45 & 38.24 & 39.11 & 36.14 & 35.47 & 36.70 \\
        \bottomrule
    \end{tabular}
    }
    \end{minipage}
    \begin{minipage}{\textwidth}
        \centering
    \resizebox{0.7\textwidth}{!}{%
    \begin{tabular}{lcccccccccc}
        \toprule
        &\multicolumn{5}{c}{Transcription Factor Prediction (Human)} & \multicolumn{3}{c}{Core Promoter Detection} \\
        \cmidrule(lr){2-6} \cmidrule(lr){7-9}
        Attack & tf0 & tf1 & tf2 & tf3 & tf4 & all & notata & tata \\ %
        \midrule
        BertAttack & \textbf{100.00} & \textbf{100.00} & \underline{99.72} & \textbf{100.00} & \textbf{100.00} & \underline{99.76} & \underline{99.55} & \underline{99.27} \\
        TextFooler & \textbf{100.00} & \textbf{100.00} & \textbf{100.00} & \textbf{100.00} & \underline{95.39} & \textbf{100.00} & \textbf{100.00} & \textbf{100.00} \\
        FIMBA & 37.33 & 41.98 & 30.99 & 20.90 & 43.04 & 33.80 & 35.23 & 42.86 \\
        PGD & 46.85 & 48.61 & 34.57 & 39.56 & 53.13 & 38.24 & 39.04 & 57.08 \\
        \bottomrule
    \end{tabular}
    }
    \end{minipage} %
    \begin{minipage}{\textwidth}
        \centering
    \resizebox{0.55\textwidth}{!}{%
    \begin{tabular}{lcccccccc}
        \toprule
        &\multicolumn{5}{c}{Transcription Factor Prediction (Mouse)} \\
        \cmidrule(lr){2-6}
        Attack & 0 & 1 & 2 & 3 & 4  \\
        \midrule
        BertAttack & \textbf{100.00} & \textbf{99.66} & \underline{99.46} & \textbf{100.00} & \textbf{100.00}  \\
        TextFooler & \textbf{100.00} & \underline{92.47} & \textbf{100.00} & \textbf{100.00} & \textbf{100.00} &  \\
        FIMBA & 35.71 & 51.06 & 39.02 & 16.36 & 28.13  \\
        PGD & 26.10 & 41.97 & 37.61 & 45.96 & 23.91 \\
        \bottomrule
    \end{tabular}
    }
    \end{minipage} %
\end{table}

\begin{table}[ht]
    \centering
    \caption{\textbf{Performance Comparison of Adversarial Attacks on NT2.} This table shows the performance of all adversarial attacks on the Nucleotide Transformer 2 (NT2) model. All results are evaluated using the Attack Success Rate (ASR) metric. The best result is highlighted in bold, while the second‐best result is underlined.}
    \begin{minipage}{\textwidth}
    \centering
    \resizebox{0.7\textwidth}{!}{%
    \begin{tabular}{lccccccc}
        \toprule
        & \multicolumn{6}{c}{Epigenetic Marks Prediction} \\
        \cmidrule(lr){2-7}
        Attack & H3 & H3K14ac & H3K36me3 & H3K4me1 & H3K4me2 & H3K4me3 \\
        \midrule
        BertAttack & \underline{98.42} & \underline{99.62} & \underline{99.91} & \underline{99.66} & \textbf{100.00} & \textbf{100.00} \\
        TextFooler & \textbf{100.00} & \textbf{100.00} & \textbf{100.00} & \textbf{100.00} & \textbf{100.00} & \textbf{100.00} \\
        FIMBA      & 27.38        & 22.08        & 34.48        & 30.26        & 23.53        & \underline{39.71} \\
        PGD        & 43.55        & 35.86        & 16.13        & 11.19        & \underline{38.99} & 11.95         \\
        \bottomrule
    \end{tabular}
    }
    \end{minipage}

    \begin{minipage}{\textwidth}
    \centering
    \resizebox{0.7\textwidth}{!}{%
    \begin{tabular}{lccccccc}
        \toprule
        & \multicolumn{4}{c}{Epigenetic Marks Prediction} & \multicolumn{3}{c}{Promoter Detection (300bp)} \\
        \cmidrule(lr){2-5} \cmidrule(lr){6-8}
        Attack & H3K79me3 & H3K9ac & H4   & H4ac & all   & notata    & tata  \\
        \midrule
        BertAttack & \textbf{100.00} & \underline{99.53} & \underline{99.45} & \textbf{100.00} & \underline{99.70} & \textbf{95.35} & \underline{99.47} \\
        TextFooler & \textbf{100.00} & \textbf{100.00}   & \textbf{100.00}   & \textbf{100.00} & \textbf{100.00}   & \underline{88.59} & \textbf{100.00}   \\
        FIMBA      & 6.02            & 62.03            & 23.08            & 25.61            & 59.60            & 9.09           & 51.58            \\
        PGD        & \underline{34.78} & 38.82            & 32.60            & \underline{38.35} & 35.34            & 32.95          & 18.03            \\
        \bottomrule
    \end{tabular}
    }
    \end{minipage}

    \begin{minipage}{\textwidth}
        \centering
    \resizebox{0.7\textwidth}{!}{%
    \begin{tabular}{lcccccccccc}
        \toprule
        &\multicolumn{5}{c}{Transcription Factor Prediction (Human)} & \multicolumn{3}{c}{Core Promoter Detection} \\
        \cmidrule(lr){2-6} \cmidrule(lr){7-9}
        Attack & tf0    & tf1    & tf2        & tf3        & tf4    & all       & notata    & tata      \\
        \midrule
        BertAttack & \textbf{100.00} & \textbf{100.00} & \textbf{100.00} & \textbf{99.83} & \textbf{100.00} & \underline{99.63} & \underline{99.31} & \textbf{99.64} \\
        TextFooler & \textbf{100.00} & \textbf{100.00} & \underline{88.84} & \underline{99.80} & \textbf{100.00} & \textbf{99.81}  & \textbf{100.00}  & 40.23        \\
        FIMBA      & 44.71         & 28.95         & 37.18         & 33.75         & 50.55  & 45.35        & 34.48        & \underline{44.79} \\
        PGD        & \underline{50.82} & \underline{65.69} & 45.11         & 36.52         & \underline{63.40} & 11.81        & 37.73        & 37.70        \\
        \bottomrule
    \end{tabular}
    }
    \end{minipage}

    \begin{minipage}{\textwidth}
        \centering
    \resizebox{0.5\textwidth}{!}{%
    \begin{tabular}{lcccccccc}
        \toprule
        &\multicolumn{5}{c}{Transcription Factor Prediction (Mouse)} \\
        \cmidrule(lr){2-6}
        Attack & 0         & 1          & 2          & 3           & 4          \\
        \midrule
        BertAttack & \textbf{100.00} & \underline{99.59} & \textbf{99.49} & \textbf{100.00} & \textbf{100.00} \\
        TextFooler & \underline{99.78} & \textbf{99.82} & \underline{95.74} & \textbf{100.00} & \underline{97.84} &  \\
        FIMBA      & 50.00        & 42.71        & 40.70        & 38.89        & 42.50        \\
        PGD        & 38.69        & 40.22        & 15.00        & \underline{41.88} & 21.56        \\
        \bottomrule
    \end{tabular}
    }
    \end{minipage}
\end{table}

\begin{table}[ht]
    \centering
    \caption{\textbf{Performance Comparison of Adversarial Attacks on GenomeOcean.} This table shows the performance of all adversarial attacks on the GenomeOcean model. All results are evaluated using the Attack Success Rate (ASR) metric. The best result is highlighted in bold, while the second-best
result is underlined.}
    \begin{minipage}{\textwidth}
    \centering
    \resizebox{0.7\textwidth}{!}{%
    \begin{tabular}{lccccccc}
        \toprule
        & \multicolumn{6}{c}{Epigenetic Marks Prediction} \\
        \cmidrule(lr){2-7}
        Attack & H3 & H3K14ac & H3K36me3 & H3K4me1 & H3K4me2 & H3K4me3 \\
        \midrule
        BertAttack & \textbf{100.00} & \underline{99.60} & \underline{99.97} & \textbf{100.00} & \underline{99.95} & \underline{99.97} \\
        TextFooler & \underline{99.78} & \textbf{100.00} & \textbf{100.00} & \textbf{100.00} & \textbf{100.00} & \textbf{100.00} \\
        FIMBA & 45.88 & 36.14 & 24.10 & 49.35 & 53.73 & 51.95 \\
        PGD & 47.74 & 42.41 & 41.11 & 48.82 & 38.28 & 45.57 \\
        \bottomrule
    \end{tabular}
    }
    \end{minipage}
   \begin{minipage}{\textwidth}
    \centering
    \resizebox{0.7\textwidth}{!}{%
    \begin{tabular}{lccccccc}
        \toprule
        & \multicolumn{4}{c}{Epigenetic Marks Prediction} & \multicolumn{3}{c}{Promoter Detection (300bp)} \\
        \cmidrule(lr){2-5} \cmidrule(lr){6-8}
        Attack & H3K79me3 & H3K9ac & H4 & H4ac & all & notata & tata \\ %
        \midrule
        BertAttack & \underline{98.75} & \textbf{100.00} & \textbf{98.18} & \underline{98.51} & \underline{99.65} & \textbf{100.00} & \underline{97.71} \\
        TextFooler & \textbf{100.00} & \textbf{100.00} & \underline{88.89} & \textbf{100.00} & \textbf{99.87} & \textbf{100.00} & \textbf{100.00} \\
        FIMBA & 43.37 & 21.52 & 35.16 & 68.67 & 59.78 & 36.36 & 28.57 \\
        PGD & 44.12 & 48.49 & 43.45 & 18.72 & 53.34 & 41.15 & 35.22 \\
        \bottomrule
    \end{tabular}
    }
    \end{minipage}
    \begin{minipage}{\textwidth}
        \centering
    \resizebox{0.7\textwidth}{!}{%
    \begin{tabular}{lcccccccccc}
        \toprule
        &\multicolumn{5}{c}{Transcription Factor Prediction (Human)} & \multicolumn{3}{c}{Core Promoter Detection} \\
        \cmidrule(lr){2-6} \cmidrule(lr){7-9}
        Attack & tf0 & tf1 & tf2 & tf3 & tf4 & all & notata & tata \\ %
        \midrule
        BertAttack & \textbf{100.00} & \textbf{100.00} & \textbf{99.89} & \underline{99.60} & \underline{99.94} & \underline{99.83} & \underline{99.91} & \underline{99.81} \\
        TextFooler & \textbf{100.00} & \textbf{100.00} & \underline{99.88} & \textbf{99.85} & \textbf{100.00} & \textbf{100.00} & \textbf{100.00} & \textbf{100.00} \\
        FIMBA & 46.91 & 31.65 & 49.37 & 39.39 & 45.88 & 42.68 & 31.33 & 38.96 \\
        PGD & 22.98 & 22.98 & 23.95 & 33.33 & 22.06 & 41.39 & 32.15 & 39.66 \\
        \bottomrule
    \end{tabular}
    }
    \end{minipage} %
    \begin{minipage}{\textwidth}
    \centering
    \resizebox{0.5\textwidth}{!}{%
    \begin{tabular}{lcccccc}
        \toprule
        &\multicolumn{5}{c}{Transcription Factor Prediction (Mouse)} \\
        \cmidrule(lr){2-6}
        Attack & 0 & 1 & 2 & 3 & 4  \\
        \midrule
        BertAttack & \textbf{100.00} & \underline{99.83} & \underline{98.95} & \underline{98.83} & \textbf{100.00}  \\
        TextFooler & \textbf{100.00} & \textbf{99.89} & \textbf{100.00} & \textbf{99.90} & \textbf{100.00}   \\
        FIMBA & 1.16 & 53.68 & 34.83 & 57.65 & 39.47  \\
        PGD & 43.36 & 23.68 & 24.94 & 32.90 & 38.91 \\
        \bottomrule
    \end{tabular}
    }
    \end{minipage} %
\end{table}

\begin{table}[ht]
    \centering
    \caption{\textbf{Performance Comparison of Adversarial Defense on DNABERT-2.} This table shows the performance of all adversarial defense on the DNABERT-2 model. All results are evaluated using the Defense Success Rate (DSR) metric. The best result is highlighted in bold, while the second-best
result is underlined.}

    \begin{minipage}{\textwidth}
    \centering
    \resizebox{0.7\textwidth}{!}{%
    \begin{tabular}{lcccccccc}
        \toprule
        & & \multicolumn{6}{c}{Epigenetic Marks Prediction} \\
        \cmidrule(lr){3-8}
        Attack & Defense & H3 & H3K14ac & H3K36me3 & H3K4me1 & H3K4me2 & H3K4me3 \\
        \midrule
        \multirow{3}{*}{PGD} &    
        FreeLB & \underline{56.17} & \underline{65.68} & \underline{66.22} & \underline{63.10} & 72.38 & \underline{63.92} \\
        & ADFAR  & \textbf{64.32} & 63.55 & 65.51 & 62.01 & \underline{74.57} & \textbf{64.58} \\
        & AT     & 54.87 & \textbf{77.97} & \textbf{69.08} & \textbf{72.55} & \textbf{82.38} & 61.01 \\
        \midrule
        \multirow{3}{*}{BertAttack} &  FreeLB & \underline{5.10} & 0.00 & \underline{1.16} & 0.00 & 1.19 & \underline{10.00} \\
        & ADFAR     & \textbf{100.00} & 0.00 & \textbf{10.10} & 0.00 & \underline{2.08} & \textbf{94.23} \\
        & AT    & 4.76 & 0.00 & 0.00 & 0.00 & \textbf{2.86} & 0.00 \\
        \midrule
        \multirow{3}{*}{TextFooler} & FreeLB & 33.88 & \underline{0.11} & 0.00 & 0.00 & 0.00 & 0.00 \\
        & ADFAR     & \textbf{42.28} & 0.00 & 0.00 & 0.00 & 0.00 & \textbf{0.22} \\
        & AT    & \underline{41.25} & \textbf{0.12} & \textbf{0.12} & 0.00 & \textbf{1.88} & 0.00 \\
        \bottomrule
    \end{tabular}
    }
    \end{minipage}

    \begin{minipage}{\textwidth}
    \centering
    \resizebox{0.7\textwidth}{!}{%
    \begin{tabular}{lcccccccc}
        \toprule
       &  & \multicolumn{4}{c}{Epigenetic Marks Prediction} & \multicolumn{3}{c}{Promoter Detection (300bp)} \\
        \cmidrule(lr){3-6} \cmidrule(lr){7-9}
       Attack & Defense & H3K79me3 & H3K9ac & H4 & H4ac & all & notata & tata \\
        \midrule
        \multirow{3}{*}{PGD} &  FreeLB & 61.47 & \textbf{63.44} & 60.84 & \textbf{67.58} & 55.93 & 56.01 & 58.74 \\
        & ADFAR  & \underline{62.08} & 55.82 & \underline{65.56} & \underline{62.38} & \textbf{70.01} & \textbf{65.59} & \textbf{64.26} \\
        & AT     & \textbf{62.91} & \underline{60.92} & \textbf{73.12} & 59.48 & \underline{63.67} & \underline{51.98} & \underline{49.74} \\
        \midrule
        \multirow{3}{*}{BertAttack} & FreeLB & 0.00 & 1.08 & \underline{6.19} & 0.00 & 0.00 & 1.00 & \textbf{9.28} \\
        & ADFAR     & 0.00 & \textbf{8.42} & 0.00 & \textbf{25.00} & \textbf{4.08} & \textbf{100.00} & 7.69 \\
        & AT    & \textbf{4.55} & \underline{4.29} & \textbf{15.62} & 0.00 & \underline{2.04} & \underline{19.59} & \underline{8.75} \\
        \midrule
        \multirow{3}{*}{TextFooler} & FreeLB & 0.00 & 0.00 & 34.68 & 0.00 & 0.00 & 3.04 & 73.16 \\
        & ADFAR     & 0.00 & 0.00 & \textbf{76.39} & \textbf{4.74} & \textbf{8.42} & \textbf{100.00} & \textbf{88.83} \\
        & AT    & \textbf{1.28} & \textbf{5.57} & \underline{38.16} & 0.00 & 0.00 & \underline{28.97} & \underline{75.63} \\
        \bottomrule
    \end{tabular}
    }
    \end{minipage}

    \begin{minipage}{\textwidth}
        \centering
    \resizebox{0.7\textwidth}{!}{%
    \begin{tabular}{lccccccccccc}
        \toprule
        & &\multicolumn{5}{c}{Transcription Factor Prediction (Human)} & \multicolumn{3}{c}{Core Promoter Detection} \\
        \cmidrule(lr){3-7} \cmidrule(lr){8-10}
        Attack & Defense & tf0 & tf1 & tf2 & tf3 & tf4 & all & notata & tata \\
        \midrule
        \multirow{3}{*}{PGD} &  FreeLB & \textbf{66.17} & \textbf{72.23} & \underline{73.21} & \textbf{66.79} & \textbf{65.54} & \underline{73.30} & \underline{69.31} & \textbf{64.18} \\
        & ADFAR  & \underline{64.78} & 64.38 & 56.85 & 56.18 & \underline{61.97} & 60.61 & 67.32 & 59.56 \\
        & AT     & 64.44 & \underline{64.76} & \textbf{77.58} & \underline{59.93} & 57.08 & \textbf{74.31} & \textbf{76.35} & \underline{62.18} \\
        \midrule
        \multirow{3}{*}{BertAttack} & FreeLB & \textbf{10.20} & 0.00 & \underline{10.00} & \textbf{2.15} & \underline{2.27} & 0.00 & 0.00 & 0.00 \\
        & ADFAR     & 0.00 & 0.00 & 0.00 & 0.00 & \textbf{100.00} & \textbf{27.08} & \textbf{7.07} & 0.00 \\
        & AT    & 0.00 & 0.00 & \textbf{10.34} & 0.00 & 0.00 & \underline{1.20} & \underline{1.14} & \textbf{1.10} \\
        \midrule
         \multirow{3}{*}{TextFooler} & FreeLB & \underline{0.22} & 0.00 & 0.00 & \underline{0.34} & 0.71 & 0.00 & \underline{1.01} & 72.85 \\
        & ADFAR     & 0.00 & 0.00 & \textbf{6.29} & \textbf{100.00} & \underline{2.41} & \textbf{26.29} & \textbf{1.61} & \textbf{97.97} \\
        & AT    & \textbf{0.98} & 0.00 & \underline{0.24} & 0.13 & \textbf{3.17} & 0.00 & 0.66 & \underline{75.18} \\
        \bottomrule
    \end{tabular}
    }
    \end{minipage}

    \begin{minipage}{\textwidth}
    \centering
    \resizebox{0.55\textwidth}{!}{%
    \begin{tabular}{lccccccc}
        \toprule
        & &\multicolumn{5}{c}{Transcription Factor Prediction (Mouse)} \\
        \cmidrule(lr){3-7}
        Attack & Defense & 0 & 1 & 2 & 3 & 4  \\
        \midrule
        \multirow{3}{*}{PGD}& FreeLB & \underline{57.93} & \underline{70.40} & 56.17 & \underline{57.29} & \textbf{61.82} \\
        & ADFAR  & \textbf{69.44} & 64.73 & \underline{60.40} & \textbf{62.41} & \underline{61.54} \\
        & AT     & 55.61 & \textbf{73.15} & \textbf{73.22} & 53.08 & 56.45 \\
        \midrule
        \multirow{3}{*}{BertAttack}& FreeLB & \underline{20.62} & 4.12 & \textbf{9.00} & \textbf{17.35} & \underline{2.20} \\
        & ADFAR  & \textbf{44.44} & \textbf{27.27} & 0.00 & \underline{10.42} & \textbf{100.00} \\
        & AT     & 5.49 & \underline{10.20} & \underline{6.82} & 5.71 & 1.10 \\
        \midrule
        \multirow{3}{*}{TextFooler}& FreeLB & 65.90 & 0.00 & 85.89 & 89.98 & 16.28 \\
        & ADFAR  & \underline{67.49} & \textbf{17.54} & \textbf{91.92} & \textbf{96.23} & \textbf{26.15} \\
        & AT     & \textbf{68.2} & \underline{6.18} & \underline{87.45} & \underline{92.45} & \underline{17.54} \\
        \bottomrule
    \end{tabular}
    }
    \end{minipage}
\end{table}

\begin{table}[ht]
    \centering
    \caption{\textbf{Performance Comparison of Adversarial Defense on GenomeOcean.} This table shows the performance of all adversarial defense on the GenomeOcean model. All results are evaluated using the Defense Success Rate (DSR) metric. The best result is highlighted in bold, while the second-best result is underlined.}

    \begin{minipage}{\textwidth}
    \centering
    \resizebox{0.7\textwidth}{!}{%
    \begin{tabular}{lcccccccc}
        \toprule
       &  & \multicolumn{6}{c}{Epigenetic Marks Prediction} \\
        \cmidrule(lr){3-8}
        Attack & Defense & H3 & H3K14ac & H3K36me3 & H3K4me1 & H3K4me2 & H3K4me3 \\
        \midrule
         \multirow{3}{*}{PGD} & FreeLB & \textbf{58.51} & 50.75 & \underline{52.96} & \underline{55.52} & 58.13 & \textbf{56.48} \\
        & ADFAR  & 54.75 & \textbf{66.59} & 49.43 & \textbf{68.20} & \textbf{69.17} & 50.24 \\
        & AT     & \underline{57.40} & \underline{55.78} & \textbf{59.35} & 49.87 & \underline{64.69} & \underline{52.15} \\
        \midrule
         \multirow{3}{*}{BertAttack} & FreeLB & \textbf{2.04} & \textbf{8.60} & \textbf{3.19} & 0.00 & 0.00 & 0.00 \\
        & ADFAR     & 0.00 & 0.00 & 0.00 & 0.00 & 0.00 & 0.00 \\
        & AT    & \underline{0.22} & \underline{4.45} & \underline{0.13} & \textbf{0.04} & \textbf{0.22} & 0.00 \\
        \midrule
      \multirow{3}{*}{TextFooler} & FreeLB & 0.00 & 0.00 & 0.00 & 0.00 & 0.00 & 0.00 \\
      &  ADFAR     & 0.00 & 0.00 & 0.00 & 0.00 & 0.00 & 0.00 \\
       & AT    & \textbf{33.75} & 0.00 & 0.00 & 0.00 & 0.00 & 0.00 \\
        \bottomrule
    \end{tabular}
    }
    \end{minipage}

    \begin{minipage}{\textwidth}
    \centering
    \resizebox{0.7\textwidth}{!}{%
    \begin{tabular}{lcccccccc}
        \toprule
        & & \multicolumn{4}{c}{Epigenetic Marks Prediction} & \multicolumn{3}{c}{Promoter Detection (300bp)} \\
        \cmidrule(lr){3-6} \cmidrule(lr){7-9}
        Attack & Defense & H3K79me3 & H3K9ac & H4 & H4ac & all & notata & tata \\
        \midrule
        \multirow{3}{*}{PGD} &  FreeLB & \textbf{57.31} & \underline{55.04} & 56.79 & \textbf{93.99} & 45.78 & \underline{52.47} & 63.83 \\
        & ADFAR  & 51.14 & 46.38 & \textbf{61.72} & 86.29 & \underline{52.74} & 51.28 & \underline{64.24} \\
        & AT     & \underline{56.04} & \textbf{55.60} & \underline{56.85} & \underline{92.58} & \textbf{53.46} & \textbf{65.77} & \textbf{66.48} \\
        \midrule
        \multirow{3}{*}{BertAttack} & FreeLB & \textbf{6.12} & \textbf{22.99} & \textbf{24.24} & \textbf{1.05} & 0.00 & 0.00 & 0.00 \\
        & ADFAR     & 0.00 & 0.00 & 0.00 & 0.00 & 0.00 & \textbf{3.77} & 0.00 \\
        & AT    & \underline{1.05} & \underline{1.69} & \underline{1.31} & \underline{0.75} & 0.00 & \underline{0.04} & 0.00 \\
        \midrule
          \multirow{3}{*}{TextFooler} & FreeLB & 0.00 & 0.00 & \underline{35.25} & 0.00 & 0.00 & 0.00 & 73.8 \\
      &  ADFAR     & 0.00 & 0.00 & 0.51 & 0.00 & 0.00 & \textbf{100.00} & \textbf{100.00} \\
      &  AT    & 0.00 & 0.00 & \textbf{37.13} & 0.00 & 0.00 & 0.00 & \underline{74.10} \\
        \bottomrule
    \end{tabular}
    }
    \end{minipage}

    \begin{minipage}{\textwidth}
        \centering
    \resizebox{0.7\textwidth}{!}{%
    \begin{tabular}{lccccccccccc}
        \toprule
        & &\multicolumn{5}{c}{Transcription Factor Prediction (Human)} & \multicolumn{3}{c}{Core Promoter Detection} \\
        \cmidrule(lr){3-7} \cmidrule(lr){8-10}
        Attack & Defense & tf0 & tf1 & tf2 & tf3 & tf4 & all & notata & tata \\
        \midrule
        \multirow{3}{*}{PGD} & FreeLB & \textbf{96.51} & 91.09 & 91.18 & \underline{67.74} & 91.79 & \textbf{70.92} & \underline{66.86} & \underline{57.31} \\
        & ADFAR  & \underline{92.09} & \textbf{97.83} & \underline{93.73} & 67.07 & \textbf{96.94} & 57.38 & 58.62 & 55.30 \\
        & AT     & 91.54 & \underline{93.95} & \textbf{94.37} & \textbf{68.14} & \underline{96.53} & \underline{60.82} & \textbf{69.29} & \textbf{61.32} \\
        \midrule
        \multirow{3}{*}{BertAttack} & FreeLB & 0.00 & 0.00 & 0.00 & 0.00 & \textbf{1.00} & \textbf{2.15} & 0.00 & \textbf{1.01} \\
        & ADFAR     & 0.00 & 0.00 & 0.00 & 0.00 & 0.00 & \underline{1.85} & 0.00 & 0.00 \\
        & AT    & 0.00 & 0.00 & 0.00 & 0.00 & 0.00 & 0.76 & \textbf{0.80} & \underline{0.18} \\
        \midrule
       \multirow{3}{*}{TextFooler}  & FreeLB & 0.00 & 0.00 & 0.00 & 0.00 & 0.00 & 0.00 & 0.00 & \underline{73.13} \\
        & ADFAR     & \textbf{100.00} & \textbf{100.00} & \textbf{100.00} & \textbf{100.00} & \textbf{100.00} & 0.00 & 0.00 & 0.10 \\
        & AT    & 0.00 & \underline{0.52} & 0.00 & 0.00 & \underline{0.42} & 0.00 & 0.00 & \textbf{74.49} \\
        \bottomrule
    \end{tabular}
    }
    \end{minipage}

    \begin{minipage}{\textwidth}
    \centering
    \resizebox{0.55\textwidth}{!}{%
    \begin{tabular}{lccccccc}
        \toprule
        & &\multicolumn{5}{c}{Transcription Factor Prediction (Mouse)} \\
        \cmidrule(lr){3-7}
      Attack &  Defense & 0 & 1 & 2 & 3 & 4  \\
        \midrule
      \multirow{3}{*}{PGD} & FreeLB & \underline{57.25} & \textbf{73.37} & \underline{68.87} & \underline{67.39} & 57.16 \\
       & ADFAR  & 55.60 & 69.74 & \textbf{69.72} & \textbf{68.53} & \underline{57.96} \\
       & AT     & \textbf{58.48} & \underline{70.22} & 48.47 & 61.68 & \textbf{58.82} \\
       \midrule
        \multirow{3}{*}{BertAttack}& FreeLB & 0.00 & \underline{1.05} & \underline{2.00} & \underline{1.00} & 0.00 \\
        & ADFAR  & 0.00 & \textbf{25.00} & 0.00 & 0.00 & 0.00 \\
        & AT     & 0.00 & 0.00 & \textbf{2.02} & \textbf{2.00} & 0.00 \\
        \midrule
        \multirow{3}{*}{TextFooler}& FreeLB & 64.44 & 0.00 & \underline{85.73} & 89.57 & \underline{28.76} \\
        & ADFAR  & \textbf{100.00} & \textbf{100.00} & \textbf{100.00} & \textbf{100.00} & \textbf{100.00} \\
        & AT     & \underline{65.98} & \underline{1.65} & 85.47 & \underline{90.03} & 17.63 \\
        \bottomrule
    \end{tabular}
    }
    \end{minipage}
\end{table}

\begin{table}[ht]
    \centering
    \caption{\textbf{Performance Comparison of Adversarial Defense on NT.} This table shows the performance of all adversarial defense on the Nucleotide Transformer (NT) model. All results are evaluated using the Defense Success Rate (DSR) metric. The best result is highlighted in bold, while the second-best result is underlined.}

    \begin{minipage}{\textwidth}
    \centering
    \resizebox{0.7\textwidth}{!}{%
    \begin{tabular}{lcccccccc}
        \toprule
        & & \multicolumn{6}{c}{Epigenetic Marks Prediction} \\
        \cmidrule(lr){3-8}
        Attack & Defense & H3 & H3K14ac & H3K36me3 & H3K4me1 & H3K4me2 & H3K4me3 \\
        \midrule
        \multirow{3}{*}{PGD} & FreeLB & \underline{87.79} & \underline{84.45} & \underline{80.44} & \underline{85.20} & \underline{84.35} & \underline{74.08} \\
        & ADFAR  & 54.65 & 53.07 & 50.08 & 57.23 & 54.72 & 57.95 \\
         & AT     & \textbf{92.35} & \textbf{86.74} & \textbf{82.02} & \textbf{86.80} & \textbf{87.54} & \textbf{75.02} \\
    \midrule
    \multirow{3}{*}{BertAttack} & FreeLB & \textbf{7.14} & 0.00 & 0.00 & \textbf{1.18} & 0.00 & 0.00 \\
       & ADFAR     & \underline{2.04} & 0.00 & 0.00 & 0.00 & \textbf{13.56} & 0.00 \\
         & AT    & 0.22 & 0.00 & 0.00 & 0.00 & 0.00 & 0.00 \\
         \midrule
         \multirow{3}{*}{TextFooler}  &  FreeLB & \underline{25.69} & 23.10 & 10.30 & \underline{12.40} & \underline{20.00} & \textbf{9.54} \\
        & ADFAR     & 0.00 & \textbf{100.00} & \textbf{62.70} & \textbf{12.90} & 9.35 & 7.33 \\
        & AT    & \textbf{47.68} & \underline{24.97} & \underline{12.31} & 9.39 & \textbf{47.97} & \underline{7.99} \\
        \bottomrule
    \end{tabular}
    }
    \end{minipage}

    \begin{minipage}{\textwidth}
    \centering
    \resizebox{0.7\textwidth}{!}{%
    \begin{tabular}{lcccccccc}
        \toprule
        & & \multicolumn{4}{c}{Epigenetic Marks Prediction} & \multicolumn{3}{c}{Promoter Detection (300bp)} \\
        \cmidrule(lr){3-6} \cmidrule(lr){7-9}
        Attack & Defense & H3K79me3 & H3K9ac & H4 & H4ac & all & notata & tata \\
        \midrule
         \multirow{3}{*}{PGD} & FreeLB & \textbf{85.64} & \textbf{83.36} & \textbf{89.65} & \textbf{84.87} & \underline{93.93} & \underline{95.41} & \textbf{99.34} \\
        & ADFAR  & 53.70 & 61.02 & 59.09 & 59.92 & 52.09 & 51.25 & 57.63 \\
        & AT     & \underline{84.74} & \underline{81.22} & \underline{82.81} & \underline{81.39} & \textbf{94.26} & \textbf{96.97} & \underline{90.45} \\
        \midrule
        \multirow{3}{*}{BertAttack} & FreeLB & 0.00 & 0.00 & \underline{2.06} & 0.00 & 0.00 & 0.00 & \underline{2.02} \\
        & ADFAR     & \textbf{6.52} & 0.00 & \textbf{2.17} & 0.00 & 0.00 & \textbf{43.75} & \textbf{11.76} \\
       & AT    & 0.00 & 0.00 & 2.04 & 0.00 & 0.00 & \underline{1.02} & 0.00 \\
        \midrule
        \multirow{3}{*}{TextFooler} &  FreeLB & \textbf{22.17} & \underline{41.03} & \underline{62.86} & \underline{35.14} & 35.79 & 31.25 & \underline{85.07} \\
        & ADFAR     & 2.55 & \textbf{100.00} & \textbf{72.48} & \textbf{42.77} & \textbf{49.20} & \textbf{69.14} & \textbf{91.32} \\
        & AT    & \underline{13.34} & 24.61 & 53.74 & 23.82 & \underline{35.97} & \underline{34.56} & 82.09 \\
        \bottomrule
    \end{tabular}
    }
    \end{minipage}

    \begin{minipage}{\textwidth}
        \centering
    \resizebox{0.7\textwidth}{!}{%
    \begin{tabular}{lccccccccccc}
        \toprule
        & &\multicolumn{5}{c}{Transcription Factor Prediction (Human)} & \multicolumn{3}{c}{Core Promoter Detection} \\
        \cmidrule(lr){3-7} \cmidrule(lr){8-10}
        Attack & Defense & tf0 & tf1 & tf2 & tf3 & tf4 & all & notata & tata \\
        \midrule
        \multirow{3}{*}{PGD} & FreeLB & 57.58 & 55.28 & \underline{72.30} & \textbf{82.95} & 48.23 & \textbf{85.07} & \underline{89.47} & \underline{39.77} \\
        & ADFAR  & \textbf{96.49} & \textbf{97.26} & \textbf{92.94} & 59.67 & \textbf{96.92} & 54.34 & 56.28 & \textbf{95.70} \\
        & AT     & \underline{84.21} & \underline{59.95} & 66.17 & \underline{62.06} & \underline{64.31} & \underline{81.73} & \textbf{91.93} & 38.15 \\
        \midrule
        \multirow{3}{*}{BertAttack} & FreeLB & 0.00 & 0.00 & 0.00 & 0.00 & 0.00 & \underline{1.04} & \textbf{1.06} & \textbf{1.02} \\
        & ADFAR     & 0.00 & 0.00 & 0.00 & \textbf{5.66} & 0.00 & 0.00 & 0.00 & 0.00 \\
        & AT    & 0.00 & 0.00 & 0.00 & 0.00 & 0.00 & \textbf{1.15} & 0.00 & \textbf{1.02} \\
        \midrule
        \multirow{3}{*}{TextFooler} & FreeLB & 36.03 & 34.17 & \underline{32.44} & 28.15 & \underline{38.83} & \underline{44.54} & \underline{47.10} & \underline{89.26} \\
        & ADFAR     & \textbf{100.00} & \textbf{60.02} & \textbf{75.00} & \textbf{99.58} & \textbf{89.74} & \textbf{64.10} & \textbf{100.00} & \textbf{100.00} \\
        & AT    & \underline{43.85} & \underline{41.76} & 28.65 & \underline{44.43} & 37.16 & 34.18 & 35.66 & 86.49 \\
        \bottomrule
    \end{tabular}
    }
    \end{minipage}

    \begin{minipage}{\textwidth}
    \centering
    \resizebox{0.55\textwidth}{!}{%
    \begin{tabular}{lcccccc}
        \toprule
      &  &\multicolumn{5}{c}{Transcription Factor Prediction (Mouse)} \\
        \cmidrule(lr){3-7}
       Attack &  Defense & 0 & 1 & 2 & 3 & 4  \\
        \midrule
       \multirow{3}{*}{PGD} & FreeLB & \underline{74.60} & \underline{98.03} & \underline{86.32} & \textbf{70.60} & \underline{75.08} \\
       & ADFAR  & 56.57 & 55.57 & 53.62 & \underline{52.30} & 59.28 \\
      &  AT     & \textbf{76.37} & \textbf{99.44} & \textbf{99.46} & 34.72 & \textbf{75.64} \\
       \midrule
        \multirow{3}{*}{BertAttack}& FreeLB & 0.00 & 0.00 & 0.00 & \textbf{2.02} & 0.00 \\
        & ADFAR  & 0.00 & \textbf{41.07} & \textbf{2.13} & 0.00 & 0.00 \\
        & AT     & 0.00 & 0.13 & 0.00 & 0.00 & 0.00 \\
        \midrule
        \multirow{3}{*}{TextFooler}& FreeLB & \underline{75.28} & \underline{58.13} & \underline{92.57} & 93.98 & \underline{31.72} \\
        & ADFAR  & \textbf{85.24} & \textbf{83.82} & \textbf{97.60} & \textbf{100.00} & \textbf{69.05} \\
        & AT     & 72.34 & 56.08 & 89.80 & \underline{94.44} & 31.63 \\
        \bottomrule
    \end{tabular}
    }
    \end{minipage}
\end{table}

\begin{table}[ht]
    \centering
    \caption{\textbf{Performance Comparison of Adversarial Defense on NT2.} This table shows the performance of all adversarial defense on the Nucleotide Transformer-2 (NT2) model. All results are evaluated using the Defense Success Rate (DSR) metric. The best result is highlighted in bold, while the second-best result is underlined.}

    \begin{minipage}{\textwidth}
    \centering
    \resizebox{0.7\textwidth}{!}{%
    \begin{tabular}{lcccccccc}
        \toprule
        & & \multicolumn{6}{c}{Epigenetic Marks Prediction} \\
        \cmidrule(lr){3-8}
       Attack & Defense & H3 & H3K14ac & H3K36me3 & H3K4me1 & H3K4me2 & H3K4me3 \\
        \midrule
       \multirow{3}{*}{PGD} & FreeLB & \underline{89.10} & \underline{80.99} & \underline{79.18} & \underline{84.75} & \textbf{76.23} & \underline{76.21} \\
        & ADFAR  & 86.57 & 73.38 & 77.85 & 77.95 & 55.52 & 67.38 \\
        & AT     & \textbf{97.61} & \textbf{82.31} & \textbf{83.20} & \textbf{86.88} & \underline{75.67} & \textbf{77.66} \\
        \midrule
        \multirow{3}{*}{BertAttack} & FreeLB & \textbf{2.02} & 0.00 & 0.00 & 0.00 & 0.00 & 0.00 \\
        & ADFAR     & 0.00 & \textbf{5.97} & \textbf{1.67} & 0.00 & 0.00 & 0.00 \\
       &  AT    & 0.00 & 0.00 & 0.00 & 0.00 & 0.00 & 0.00 \\
       \midrule
        \multirow{3}{*}{TextFooler} & FreeLB & 33.23          & 0.00 & 0.00 & 0.00 & 0.00 & 0.00 \\
       & ADFAR  & \textbf{49.57} & 0.00 & 0.00 & 0.00 & 0.00 & 0.00 \\
        & AT     & \underline{35.70} & 0.00 & 0.00 & 0.00 & 0.00 & 0.00 \\
        \bottomrule
    \end{tabular}
    }
    \end{minipage}

    \begin{minipage}{\textwidth}
    \centering
    \resizebox{0.7\textwidth}{!}{%
    \begin{tabular}{lcccccccc}
        \toprule
        & & \multicolumn{4}{c}{Epigenetic Marks Prediction} & \multicolumn{3}{c}{Promoter Detection (300bp)} \\
        \cmidrule(lr){3-6} \cmidrule(lr){7-9}
        Attack & Defense & H3K79me3 & H3K9ac & H4 & H4ac & all & notata & tata \\
        \midrule
        \multirow{3}{*}{PGD} & FreeLB & \textbf{89.83} & \underline{84.55} & \textbf{99.44} & \underline{79.34} & \textbf{94.93} & \textbf{91.57} & \textbf{94.00} \\
        & ADFAR  & 89.28 & 73.77 & 74.60 & 73.34 & 61.27 & 57.61 & 70.18 \\
       &  AT     & \underline{89.43} & \textbf{86.33} & \underline{96.76} & \textbf{83.78} & \underline{93.74} & \underline{90.42} & \underline{85.16} \\
       \midrule
       \multirow{3}{*}{BertAttack} &  FreeLB & 0.00 & 0.00 & \underline{3.12} & 0.00 & 0.00 & 0.00 & \textbf{1.00} \\
       &  ADFAR     & \textbf{18.18} & 0.00 & \textbf{4.26} & 0.00 & 0.00 & \textbf{18.75} & 0.00 \\
        & AT    & 0.00 & 0.00 & 1.02 & 0.00 & 0.00 & 0.00 & 0.00 \\
        \midrule
       \multirow{3}{*}{TextFooler} &  FreeLB & 0.00 & \textbf{0.11} & \underline{35.29} & 0.00 & \textbf{0.71} & 0.00          & 73.73 \\
        & ADFAR  & 0.00 & 0.00          & \textbf{72.82} & 0.00 & 0.00          & \textbf{0.71} & \textbf{76.05} \\
        & AT     & 0.00 & 0.00          & 35.22          & 0.00 & 0.00          & 0.00          & \underline{74.33} \\
        \bottomrule
    \end{tabular}
    }
    \end{minipage}

    \begin{minipage}{\textwidth}
        \centering
    \resizebox{0.7\textwidth}{!}{%
    \begin{tabular}{lccccccccccc}
        \toprule
        & &\multicolumn{5}{c}{Transcription Factor Prediction (Human)} & \multicolumn{3}{c}{Core Promoter Detection} \\
        \cmidrule(lr){3-7} \cmidrule(lr){8-10}
        Attack & Defense & tf0 & tf1 & tf2 & tf3 & tf4 & all & notata & tata \\
        \midrule
        \multirow{3}{*}{PGD} &  FreeLB & \textbf{92.86} & \textbf{94.01} & \textbf{82.76} & \underline{84.25} & \textbf{97.22} & \textbf{91.26} & \textbf{91.33} & \textbf{99.82} \\
        & ADFAR  & \underline{73.62} & \underline{68.87} & \underline{73.46} & 71.17 & 75.97 & 73.68 & \underline{78.40} & \underline{62.35} \\
       & AT     & 61.98 & \underline{68.87} & 61.98 & \textbf{87.24} & \underline{94.12} & \underline{88.43} & 76.66 & 55.54 \\
       \midrule
     \multirow{3}{*}{BertAttack}  & FreeLB & 0.00 & 0.00 & 0.00 & \textbf{2.22} & 0.00 & 0.00 & 0.00 & 0.00 \\
        & ADFAR     & \textbf{51.06} & \textbf{60.38} & 0.00 & 0.00 & \textbf{2.04} & 0.00 & 0.00 & 0.00 \\
       &  AT    & 0.00 & \underline{4.00} & \textbf{1.00} & 0.00 & 0.00 & 0.00 & 0.00 & \textbf{1.00} \\
       \midrule
       \multirow{3}{*}{TextFooler} & FreeLB & 0.00 & 0.00 & 0.00          & 0.00          & 0.00 & 0.00 & 0.00 & 72.76          \\
       & ADFAR  & 0.00 & 0.00 & 0.00          & 0.00          & 0.00 & 0.00 & 0.00 & \textbf{85.48} \\
        & AT     & 0.00 & 0.00 & \textbf{0.15} & \textbf{0.12} & 0.00 & 0.00 & 0.00 & \underline{77.81} \\
        \bottomrule
    \end{tabular}
    }
    \end{minipage}

    \begin{minipage}{\textwidth}
    \centering
    \resizebox{0.55\textwidth}{!}{%
    \begin{tabular}{lcccccc}
        \toprule
       & &\multicolumn{5}{c}{Transcription Factor Prediction (Mouse)} \\
        \cmidrule(lr){3-7}
       Attack & Defense & 0 & 1 & 2 & 3 & 4  \\
        \midrule
        \multirow{3}{*}{PGD} & FreeLB & \textbf{88.71} & \textbf{99.19} & 97.29 & \underline{81.65} & \textbf{81.44} \\
        & ADFAR  & \underline{77.22} & 74.06 & \textbf{99.56} & 66.95 & \underline{61.05} \\
        & AT     & 74.61 & \underline{97.07} & \textbf{99.56} & \textbf{86.09} & 51.29 \\
        \midrule
        \multirow{3}{*}{BertAttack}& FreeLB & 0.00 & \textbf{4.04} & \textbf{2.00} & \textbf{4.08} & 0.00 \\
        & ADFAR  & \textbf{1.92} & 0.00 & 0.00 & 0.00 & \textbf{16.67} \\
        & AT     & 0.00 & \underline{4.00} & \underline{1.00} & 0.00 & 0.00 \\
        \midrule
        \multirow{3}{*}{TextFooler}& FreeLB & 63.98 & 0.00 & 85.96 & 89.66 & 16.67 \\
        & ADFAR  & \textbf{77.00} & 0.00 & \underline{86.34} & \textbf{94.90} & \textbf{29.07} \\
        & AT     & \underline{67.30} & 0.20 & \textbf{86.69} & \underline{92.44} & \underline{22.71} \\
        \bottomrule
    \end{tabular}
    }
    \end{minipage}
\end{table}

\begin{table}[ht]
    \centering
   \caption{\textbf{Performance Comparison of Adversarial Defense on HyenaDNA.} This table shows the performance of all adversarial defense on the HyenaDNA model. All results are evaluated using the Defense Success Rate (DSR) metric. The best result is highlighted in bold, while the second-best result is underlined.}

    \begin{minipage}{\textwidth}
    \centering
    \resizebox{0.7\textwidth}{!}{%
    \begin{tabular}{lcccccccc}
        \toprule
       &  & \multicolumn{6}{c}{Epigenetic Marks Prediction} \\
        \cmidrule(lr){3-8}
        Attack & Defense & H3 & H3K14ac & H3K36me3 & H3K4me1 & H3K4me2 & H3K4me3 \\
        \midrule
        \multirow{3}{*}{PGD} & FreeLB & 76.72 & 70.87 & \underline{98.19} & \underline{91.86} & \underline{96.22} & \underline{85.29} \\
         & ADFAR  & \textbf{88.44} & \underline{74.31} & 85.63 & \textbf{94.41} & \textbf{98.83} & 84.20 \\
        & AT     & \textbf{88.44} & \textbf{84.26} & \textbf{99.36} & 86.77 & 91.96 & \textbf{87.48} \\
        \midrule
        \multirow{3}{*}{BertAttack} & FreeLB & 0.00 & 0.00 & 0.00 & 0.00 & 0.00 & 0.00 \\
        & ADFAR     & 0.00 & 0.00 & 0.00 & 0.00 & 0.00 & 0.00 \\
        & AT    & 0.00 & 0.00 & 0.00 & 0.00 & 0.00 & 0.00 \\
        \midrule
         \multirow{3}{*}{TextFooler} & FreeLB & \textbf{100.00} & \underline{98.08} & \underline{71.00} & \textbf{75.21} & \underline{53.82} & \textbf{100.00} \\
       & ADFAR     & \textbf{100.00} & \textbf{99.77} & 30.70 & 50.62 & 29.01 & \underline{97.75} \\
        & AT    & \textbf{100.00} & 84.18 & \textbf{95.87} & \underline{50.68} & \textbf{64.87} & 80.81 \\
        \bottomrule
    \end{tabular}
    }
    \end{minipage}

    \begin{minipage}{\textwidth}
    \centering
    \resizebox{0.7\textwidth}{!}{%
    \begin{tabular}{lcccccccc}
        \toprule
       &  & \multicolumn{4}{c}{Epigenetic Marks Prediction} & \multicolumn{3}{c}{Promoter Detection (300bp)} \\
        \cmidrule(lr){3-6} \cmidrule(lr){7-9}
        Attack & Defense & H3K79me3 & H3K9ac & H4 & H4ac & all & notata & tata \\
        \midrule
        \multirow{3}{*}{PGD}  &  FreeLB & \underline{95.32} & 90.09 & \underline{62.33} & \underline{85.31} & 56.04 & \textbf{94.81} & \textbf{97.27} \\
       &  ADFAR  & 93.53 & \textbf{98.33} & 60.58 & \textbf{95.96} & \underline{83.52} & 40.20 & \underline{89.77} \\
        & AT     & \textbf{96.32} & \underline{93.99} & \textbf{63.34} & \underline{85.31} & \textbf{98.47} & \underline{49.07} & 76.80 \\
                \midrule
         \multirow{3}{*}{BertAttack} & FreeLB & 0.00 & 0.00 & 0.00 & 0.00 & 0.00 & \textbf{16.33} & 0.00 \\
        & ADFAR     & 0.00 & 0.00 & 0.00 & 0.00 & 0.00 & 0.00 & 0.00 \\
        & AT    & 0.00 & 0.00 & 0.00 & 0.00 & 0.00 & \underline{10.00} & 0.00 \\
        \midrule
         \multirow{3}{*}{TextFooler}& FreeLB & 20.42 & \underline{17.94} & \underline{90.50} & 3.28 & \underline{76.54} & \textbf{100.00} & \underline{93.46} \\
        & ADFAR     & \underline{63.24} & 15.85 & 88.98 & \underline{81.68} & 65.48 & \underline{92.86} & 89.93 \\
        & AT    & \textbf{99.64} & \textbf{45.01} & \textbf{92.80} & \textbf{85.59} & \textbf{100.00} & 27.44 & \textbf{93.97} \\
        \bottomrule
    \end{tabular}
    }
    \end{minipage}

    \begin{minipage}{\textwidth}
    \centering
    \resizebox{0.7\textwidth}{!}{%
    \begin{tabular}{lccccccccccc}
        \toprule
        & &\multicolumn{5}{c}{Transcription Factor Prediction (Human)} & \multicolumn{3}{c}{Core Promoter Detection} \\
        \cmidrule(lr){3-7} \cmidrule(lr){8-10}
       Attack &  Defense & tf0 & tf1 & tf2 & tf3 & tf4 & all & notata & tata \\
        \midrule
        \multirow{3}{*}{PGD}& FreeLB & \textbf{87.44} & \underline{87.44} & \underline{88.44} & \underline{87.44} & \textbf{88.44} & \textbf{98.47} & 85.47 & \textbf{96.26} \\
        & ADFAR  & 83.42 & \textbf{99.50} & 76.38 & \textbf{95.48} & \underline{87.44} & 68.94 & \textbf{98.61} & \underline{90.77} \\
        & AT     & \textbf{87.44} & \underline{87.44} & \textbf{91.46} & \underline{87.44} & 79.40 & \underline{96.10} & \textbf{98.61} & 83.30 \\
        \midrule
        \multirow{3}{*}{BertAttack}&  FreeLB & \underline{2.13} & 0.00 & \underline{2.04} & 0.00 & 0.00 & \textbf{6.82} & 0.00 & \textbf{1.92} \\
        & ADFAR     & 0.00 & 0.00 & 0.00 & 0.00 & 0.00 & \underline{1.85} & 0.00 & 0.00 \\
        & AT    & \textbf{5.98} & 0.00 & \textbf{3.72} & 0.00 & 0.00 & 0.00 & 0.00 & 0.00 \\
        \midrule
       \multirow{3}{*}{TextFooler} &  FreeLB & 23.33 & 19.42 & \underline{95.63} & \textbf{100.00} & 14.08 & 66.42 & 99.38 & 94.70 \\
        & ADFAR     & \textbf{100.00} & \textbf{100.00} & 89.68 & 87.00 & \textbf{100.00} & \underline{93.89} & \textbf{100.00} & \textbf{100.00} \\
       &  AT    & \textbf{100.00} & \textbf{100.00} & \textbf{100.00} & \textbf{100.00} & \textbf{100.00} & \textbf{100.00} & \textbf{100.00} & \textbf{100.00} \\
        \bottomrule
    \end{tabular}
    }
    \end{minipage}

    \begin{minipage}{\textwidth}
    \centering
    \resizebox{0.55\textwidth}{!}{%
    \begin{tabular}{lcccccc}
        \toprule
       & &\multicolumn{5}{c}{Transcription Factor Prediction (Mouse)} \\
        \cmidrule(lr){3-7}
     Attack &   Defense & 0 & 1 & 2 & 3 & 4  \\
        \midrule
       \multirow{3}{*}{PGD} & FreeLB & \textbf{94.47} & \textbf{98.59} & \textbf{75.38} & \textbf{83.76} & \underline{89.81} \\
      &  ADFAR  & 85.43 & 87.08 & \underline{62.81} & \underline{65.99} & 87.68 \\
        & AT     & \textbf{94.47} & \underline{97.23} & \underline{62.81} & \underline{65.99} & \textbf{94.09} \\
        \midrule
        \multirow{3}{*}{BertAttack}& FreeLB & 0.00 & 0.00 & 0.00 & 0.00 & 0.00 \\
        & ADFAR  & \textbf{37.04} & 0.00 & 0.00 & 0.00 & 0.00 \\
        & AT     & \underline{1.23} & 0.00 & 0.00 & 0.00 & 0.00 \\
        \midrule
        \multirow{3}{*}{TextFooler}& FreeLB & \textbf{100.00} & 19.69 & \textbf{100.00} & 94.94 & \textbf{80.78} \\
        & ADFAR  & \textbf{100.00} & \textbf{89.64} & \textbf{100.00} & \textbf{100.00} & \underline{35.34} \\
        & AT     & \textbf{100.00} & \underline{37.31} & \textbf{100.00} & \textbf{100.00} & 31.02 \\
        \bottomrule
    \end{tabular}
    }
    \end{minipage}
\end{table}

\begin{table}[ht]
    \centering
    \caption{\textbf{Performance Comparison of Adversarial Attack on Quantization Model.} This table reports the Attack Success Rate (ASR) of two adversarial attacks (TextFooler and BERTAttack) on quantized versions (Vanilla and $\text{Softmax}_1$) of DNABERT-2 and Nucleotide Transformer (NT) under W8A8 (8-bit weights and activations) quantization. All results are evaluated using the Attack Success Rate (ASR) metric.}
    \begin{minipage}{\textwidth}
    \centering
    \resizebox{\textwidth}{!}{%
    \begin{tabular}{lcccccccc}
        \toprule
        & & & \multicolumn{6}{c}{Epigenetic Marks Prediction} \\
        \cmidrule(lr){4-9}
        Attack  & Model & Quant\_Method  & H3    & H3K14ac        & H3K36me3       & H3K4me1        & H3K4me2        & H3K4me3 \\
        \midrule
        \multirow{4}{*}{TextFooler}
          & \multirow{2}{*}{DNABERT2}
              & Vanilla        & \textbf{0.19}   & \textbf{9.76}   & \textbf{24.12}  & \textbf{5.52}   & 25.53  & \textbf{12.24}   \\
          & 
              & $\text{Softmax}_1$ & 0.00   & 3.82   & 15.67  & 2.03   & \textbf{31.90}  & 4.14    \\
        \cmidrule(lr){2-9}
          & \multirow{2}{*}{NT1}
              & Vanilla        & 70.49  & \textbf{79.37}  & \textbf{77.74}  & \textbf{77.04}  & \textbf{70.49}  & \textbf{87.14}   \\
          & 
              & $\text{Softmax}_1$ &    \textbf{73.96}  & 73.65  & 77.53  & 70.89  & 70.33  & 86.21   \\
        \midrule
        \multirow{4}{*}{BertAttack}
          & \multirow{2}{*}{DNABERT2}
              & Vanilla        & \textbf{62.50}  & 26.09  & \textbf{100.00} & 61.54  & 81.25  & \textbf{100.00} \\
          &
              & $\text{Softmax}_1$ & \textbf{62.50}  & \textbf{100.00} & 16.00  & \textbf{100.00} & \textbf{93.75}  & 60.00   \\
        \cmidrule(lr){2-9}
          & \multirow{2}{*}{NT1}
              & Vanilla        & \textbf{100.00} & \textbf{100.00} & \textbf{100.00} & \textbf{100.00} & \textbf{100.00} & \textbf{100.00} \\
          &
              & $\text{Softmax}_1$ & 92.31  & \textbf{100.00} & \textbf{100.00} & \textbf{100.00} & \textbf{100.00} & 99.60   \\
        \bottomrule
    \end{tabular}
    }
    \end{minipage}

   \begin{minipage}{\textwidth}
    \centering
    \resizebox{\textwidth}{!}{%
    \begin{tabular}{lccccccccc}
        \toprule
       && & \multicolumn{4}{c}{Epigenetic Marks Prediction} & \multicolumn{3}{c}{Promoter Detection (300bp)} \\
        \cmidrule(lr){4-7} \cmidrule(lr){8-10}
        Attack & Model & Quant\_Method    & H3K79me3       & H3K9ac         & H4             & H4ac           & all            & notata         & tata \\
        \midrule
        \multirow{4}{*}{TextFooler}
          & \multirow{2}{*}{DNABERT2}
              & Vanilla        & \textbf{4.30}   & 0.00   & \textbf{11.48}  & 1.30   & 27.58  & 17.05  & \textbf{30.29}  \\
          &
              & $\text{Softmax}_1$ & 3.96   & 0.00   & 4.19   & \textbf{1.48}   & \textbf{28.21}  & \textbf{22.44}  & 29.55  \\
        \cmidrule(lr){2-10}
          & \multirow{2}{*}{NT1}
              & Vanilla        & \textbf{71.49}  & \textbf{73.37}  & \textbf{56.52}  & \textbf{72.17}  & 59.54  & 54.59  & \textbf{58.15}  \\
          &
              & $\text{Softmax}_1$ & 68.89  & 67.25  & 55.12  & 71.90  & \textbf{68.42}  & \textbf{63.40}  & \textbf{58.15}  \\
        \midrule
        \multirow{4}{*}{BertAttack}
          & \multirow{2}{*}{DNABERT2}
              & Vanilla        & \textbf{100.00} & \textbf{100.00} & \textbf{57.14}  & \textbf{99.78}  & \textbf{98.08}  & \textbf{96.43}  & 72.56  \\
          &
              & $\text{Softmax}_1$ & 84.62  & 87.50  & 0.00   & 96.15  & 66.11  & 70.00  & \textbf{100.00} \\
        \cmidrule(lr){2-10}
          & \multirow{2}{*}{NT1}
              & Vanilla        & \textbf{100.00} & \textbf{100.00} & 91.67  & \textbf{100.00} & 98.25  & 93.75  & \textbf{100.00} \\
          &
              & $\text{Softmax}_1$ & \textbf{100.00} & \textbf{100.00} & \textbf{99.27} & \textbf{100.00} & \textbf{100.00} & \textbf{97.83} & \textbf{100.00} \\
        \bottomrule
    \end{tabular}%
    }
    \end{minipage}
\end{table}

\begin{table}[ht]
    \centering
    \caption{\textbf{Performance of Adversarial Attacks on HyenaDNA Trained with the GenoAdv Dataset.} This table compares the performance of HyenDNA trained with adversarial examples from the GenoAdv dataset. Three attack methods (BERTAttack, TextFooler, and PGD) are used to evaluate the models, with results reported in terms of Attack Success Rate (ASR). The best result is highlighted in bold, while the second-best result is underlined.}

    \begin{minipage}{\textwidth}
    \centering
    \resizebox{0.7\textwidth}{!}{%
    \begin{tabular}{lccccccc}
        \toprule
        & \multicolumn{6}{c}{Epigenetic Marks Prediction} \\
        \cmidrule(lr){2-7}
        Attack & H3 & H3K14ac & H3K36me3 & H3K4me1 & H3K4me2 & H3K4me3 \\
        \midrule
        TextFooler  & 1.01 & 5.41 & \underline{83.24} & \underline{3.18} & \underline{17.86} & \underline{62.82} \\
        PGD         & \underline{12.83} & \underline{19.29} & 17.20 & 2.85 & 4.73 & 6.13 \\
        BERT\_Attack & \textbf{100.00} & \textbf{100.00} & \textbf{100.00} & \textbf{100.00} & \textbf{100.00} & \textbf{100.00} \\
        \bottomrule
    \end{tabular}%
    }
    \end{minipage}
\end{table}

\begin{table}[ht]
    \centering
    \caption{\textbf{Performance of Adversarial Attacks on GenomeOcean Trained with the GenoAdv Dataset.} This table compares the performance of GenomeOcean trained with adversarial examples from the GenoAdv dataset. Three attack methods (BERTAttack, TextFooler, and PGD) are used to evaluate the models, with results reported in terms of Attack Success Rate (ASR). The best result is highlighted in bold, while the second-best result is underlined.}

    \begin{minipage}{\textwidth}
    \centering
    \resizebox{0.7\textwidth}{!}{%
    \begin{tabular}{lccccccc}
        \toprule
        & \multicolumn{6}{c}{Epigenetic Marks Prediction} \\
        \cmidrule(lr){2-7}
        Attack & H3 & H3K14ac & H3K36me3 & H3K4me1 & H3K4me2 & H3K4me3 \\
        \midrule
        TextFooler  & \underline{62.66} & \textbf{100.00} & \textbf{100.00} & \textbf{100.00} & \textbf{100.00} & \textbf{100.00} \\
        PGD         & 34.44 & 35.87 & 24.51 & 40.00 & 39.43 & 1.36 \\
        BERT\_Attack & \textbf{100.00} & \underline{98.56} & \underline{97.65} & \textbf{100.00} & \textbf{100.00} & \textbf{100.00} \\
        \bottomrule
    \end{tabular}
    }
    \end{minipage}

    \begin{minipage}{\textwidth}
    \centering
    \resizebox{0.7\textwidth}{!}{%
    \begin{tabular}{lccccccc}
        \toprule
        & \multicolumn{4}{c}{Epigenetic Marks Prediction} & \multicolumn{3}{c}{Promoter Detection (300bp)} \\
        \cmidrule(lr){2-5} \cmidrule(lr){6-8}
        Attack & H3K79me3 & H3K9ac & H4 & H4ac & all & notata & tata \\
        \midrule
        TextFooler  & \textbf{100.00} & \textbf{100.00} & \underline{63.89} & \textbf{100.00} & \textbf{100.00} & \textbf{100.00} & 22.65 \\
        PGD         & 39.52 & 36.69 & 26.34 & 34.64 & 33.45 & 34.76 & \underline{30.91} \\
        BERT\_Attack & \underline{95.70} & \textbf{100.00} & \textbf{97.94} & \underline{98.77} & \textbf{100.00} & \underline{96.45} & \textbf{100.00} \\
        \bottomrule
    \end{tabular}
    }
    \end{minipage}

    \begin{minipage}{\textwidth}
    \centering
    \resizebox{0.7\textwidth}{!}{%
    \begin{tabular}{lcccccccccc}
        \toprule
        & \multicolumn{5}{c}{Transcription Factor Prediction (Human)} & \multicolumn{3}{c}{Core Promoter Detection} \\
        \cmidrule(lr){2-6} \cmidrule(lr){7-9}
        Attack & tf0 & tf1 & tf2 & tf3 & tf4 & all & notata & tata \\
        \midrule
        TextFooler  & \textbf{100.00} & \textbf{100.00} & \textbf{100.00} & \textbf{100.00} & \underline{99.89} & \underline{98.32} & \textbf{100.00} & 22.71 \\
        PGD         & 34.18 & 12.68 & 35.80 & 19.15 & 35.65 & 44.22 & 40.89 & \underline{39.07} \\
        BERT\_Attack & \underline{98.12} & \textbf{100.00} & \textbf{100.00} & \textbf{100.00} & \textbf{100.00} & \textbf{98.84} & \textbf{100.00} & \textbf{100.00} \\
        \bottomrule
    \end{tabular}
    }
    \end{minipage}

    \begin{minipage}{\textwidth}
    \centering
    \resizebox{0.5\textwidth}{!}{%
    \begin{tabular}{lccccc}
        \toprule
        & \multicolumn{5}{c}{Transcription Factor Prediction (Mouse)} \\
        \cmidrule(lr){2-6}
        Attack & 0 & 1 & 2 & 3 & 4 \\
        \midrule
        TextFooler  & 24.73 & \underline{96.33} & 13.58 & 8.88 & \underline{80.71} \\
        PGD         & \underline{35.06} & 30.33 & \underline{34.42} & \underline{26.60} & 25.45 \\
        BERT\_Attack & \textbf{100.00} & \textbf{100.00} & \textbf{98.96} & \textbf{100.00} & \textbf{100.00} \\
        \bottomrule
    \end{tabular}
    }
    \end{minipage}
\end{table}

\begin{table}[ht]
    \centering
   \caption{\textbf{Performance of Adversarial Attacks on DNABERT-2 Trained with the GenoAdv Dataset.} This table compares the performance of DNABERT-2 trained with adversarial examples from the GenoAdv dataset. Three attack methods (BERTAttack, TextFooler, and PGD) are used to evaluate the models, with results reported in terms of Attack Success Rate (ASR). The best result is highlighted in bold, while the second-best result is underlined.}
    \begin{minipage}{\textwidth}
    \centering
    \resizebox{0.7\textwidth}{!}{%
    \begin{tabular}{lccccccc}
        \toprule
        & \multicolumn{6}{c}{Epigenetic Marks Prediction} \\
        \cmidrule(lr){2-7}
        Attack & H3 & H3K14ac & H3K36me3 & H3K4me1 & H3K4me2 & H3K4me3 \\
        \midrule
        TextFooler  & \underline{61.83} & \textbf{100.00} & \textbf{100.00} & \textbf{100.00} & \textbf{100.00} & \textbf{100.00} \\
        PGD         & 39.53 & 24.67 & 34.53 & 36.71 & 35.61 & 34.79 \\
        BERT\_Attack & \textbf{87.67} & \underline{85.36} & \textbf{100.00} & \underline{88.63} & \underline{88.13} & \textbf{100.00} \\
        \bottomrule
    \end{tabular}
    }
    \end{minipage}

    \begin{minipage}{\textwidth}
    \centering
    \resizebox{0.7\textwidth}{!}{%
    \begin{tabular}{lccccccc}
        \toprule
        & \multicolumn{4}{c}{Epigenetic Marks Prediction} & \multicolumn{3}{c}{Promoter Detection (300bp)} \\
        \cmidrule(lr){2-5} \cmidrule(lr){6-8}
        Attack & H3K79me3 & H3K9ac & H4 & H4ac & all & notata & tata \\
        \midrule
        TextFooler  & \textbf{99.88} & \underline{69.87} & \underline{61.00} & \textbf{100.00} & 56.26 & \textbf{100.00} & 24.27 \\
        PGD         & 41.24 & 29.06 & 26.35 & 37.59 & 38.23 & 45.11 & \underline{44.93} \\
        BERT\_Attack & \underline{88.90} & \textbf{100.00} & \textbf{87.10} & \textbf{100.00} & \textbf{100.00} & \underline{88.99} & \textbf{87.56} \\
        \bottomrule
    \end{tabular}
    }
    \end{minipage}

    \begin{minipage}{\textwidth}
    \centering
    \resizebox{0.7\textwidth}{!}{%
    \begin{tabular}{lcccccccccc}
        \toprule
        & \multicolumn{5}{c}{Transcription Factor Prediction (Human)} & \multicolumn{3}{c}{Core Promoter Detection} \\
        \cmidrule(lr){2-6} \cmidrule(lr){7-9}
        Attack & tf0 & tf1 & tf2 & tf3 & tf4 & all & notata & tata \\
        \midrule
        TextFooler  & \textbf{100.00} & \underline{99.87} & \textbf{100.00} & \textbf{100.00} & \textbf{99.21} & \textbf{100.00} & \textbf{100.00} & 23.39 \\
        PGD         & 30.12 & 25.33 & 24.39 & 2.22 & 28.09 & 36.36 & 22.71 & \underline{36.89} \\
        BERT\_Attack & \underline{95.60} & \textbf{100.00} & \textbf{100.00} & \underline{97.78} & \underline{98.88} & \textbf{100.00} & \underline{98.80} & \textbf{100.00} \\
        \bottomrule
    \end{tabular}
    }
    \end{minipage}

    \begin{minipage}{\textwidth}
    \centering
    \resizebox{0.5\textwidth}{!}{%
    \begin{tabular}{lccccc}
        \toprule
        & \multicolumn{5}{c}{Transcription Factor Prediction (Mouse)} \\
        \cmidrule(lr){2-6}
        Attack & 0 & 1 & 2 & 3 & 4 \\
        \midrule
        TextFooler  & 28.54 & \textbf{98.28} & \underline{12.77} & 6.49 & \underline{81.43} \\
        PGD         & \underline{35.81} & 30.25 & 9.64 & \underline{13.00} & 34.63 \\
        BERT\_Attack & \textbf{100.00} & \underline{87.94} & \textbf{87.59} & \textbf{96.61} & \textbf{100.00} \\
        \bottomrule
    \end{tabular}
    }
    \end{minipage}
\end{table}

\begin{table}[ht]
    \centering
  \caption{\textbf{Performance of Adversarial Attacks on NT Trained with the GenoAdv Dataset.} This table compares the performance of Nucleotide Transformers (NT) trained with adversarial examples from the GenoAdv dataset. Three attack methods (BERTAttack, TextFooler, and PGD) are used to evaluate the models, with results reported in terms of Attack Success Rate (ASR). The best result is highlighted in bold, while the second-best result is underlined.}

    \begin{minipage}{\textwidth}
    \centering
    \resizebox{0.7\textwidth}{!}{%
    \begin{tabular}{lccccccc}
        \toprule
        & \multicolumn{6}{c}{Epigenetic Marks Prediction} \\
        \cmidrule(lr){2-7}
        Attack & H3 & H3K14ac & H3K36me3 & H3K4me1 & H3K4me2 & H3K4me3 \\
        \midrule
        TextFooler  & \underline{56.41} & \underline{70.39} & \underline{77.72} & \underline{85.08} & \underline{77.87} & \underline{80.64} \\
        PGD         & 28.57 & 23.43 & 21.88 & 29.53 & 21.67 & 22.90 \\
        BERT\_Attack & \textbf{100.00} & \textbf{100.00} & \textbf{100.00} & \textbf{100.00} & \textbf{100.00} & \textbf{100.00} \\
        \bottomrule
    \end{tabular}
    }
    \end{minipage}

    \begin{minipage}{\textwidth}
    \centering
    \vspace{1.5mm}
    \resizebox{0.7\textwidth}{!}{%
    \begin{tabular}{lccccccc}
        \toprule
        & \multicolumn{4}{c}{Epigenetic Marks Prediction} & \multicolumn{3}{c}{Promoter Detection (300bp)} \\
        \cmidrule(lr){2-5} \cmidrule(lr){6-8}
        Attack & H3K79me3 & H3K9ac & H4 & H4ac & all & notata & tata \\
        \midrule
        TextFooler  & \underline{79.42} & \underline{69.67} & \underline{52.19} & \underline{66.39} & \underline{46.25} & \underline{64.64} & \underline{21.50} \\
        PGD         & 17.64 & 26.87 & 7.49 & 19.89 & 19.39 & 7.97 & 7.83 \\
        BERT\_Attack & \textbf{100.00} & \textbf{100.00} & \textbf{100.00} & \textbf{100.00} & \textbf{100.00} & \textbf{100.00} & \textbf{100.00} \\
        \bottomrule
    \end{tabular}
    }
    \end{minipage}

    \begin{minipage}{\textwidth}
    \centering
    \vspace{1.5mm}
    \resizebox{0.7\textwidth}{!}{%
    \begin{tabular}{lcccccccccc}
        \toprule
        & \multicolumn{5}{c}{Transcription Factor Prediction (Human)} & \multicolumn{3}{c}{Core Promoter Detection} \\
        \cmidrule(lr){2-6} \cmidrule(lr){7-9}
        Attack & tf0 & tf1 & tf2 & tf3 & tf4 & all & notata & tata \\
        \midrule
        TextFooler  & \underline{58.31} & \underline{61.81} & \underline{46.13} & \underline{60.44} & \underline{67.96} & \underline{44.69} & \underline{67.92} & \underline{13.82} \\
        PGD         & 28.57 & 24.15 & 21.57 & 25.48 & 10.11 & 23.01 & 25.96 & 13.01 \\
        BERT\_Attack & \textbf{100.00} & \textbf{85.37} & \textbf{100.00} & \textbf{97.85} & \textbf{98.88} & \textbf{100.00} & \textbf{100.00} & \textbf{100.00} \\
        \bottomrule
    \end{tabular}
    }
    \end{minipage}

    \begin{minipage}{\textwidth}
    \centering
    \vspace{1.5mm}
    \resizebox{0.5\textwidth}{!}{%
    \begin{tabular}{lccccc}
        \toprule
        & \multicolumn{5}{c}{Transcription Factor Prediction (Mouse)} \\
        \cmidrule(lr){2-6}
        Attack & 0 & 1 & 2 & 3 & 4 \\
        \midrule
        TextFooler  & 24.55 & \underline{76.23} & 10.08 & 8.26 & \underline{66.19} \\
        PGD         & \underline{25.00} & 21.96 & \underline{10.71} & \underline{26.81} & 26.46 \\
        BERT\_Attack & \textbf{100.00} & \textbf{100.00} & \textbf{100.00} & \textbf{100.00} & \textbf{100.00} \\
        \bottomrule
    \end{tabular}
    }
    \end{minipage}
\end{table}

\begin{table}[ht]
    \centering
    \caption{\textbf{Performance of Adversarial Attacks on NT2 Trained with the GenoAdv Dataset.} This table compares the performance of Nucleotide Transformers-2 (NT2) trained with adversarial examples from the GenoAdv dataset. Three attack methods (BERTAttack, TextFooler, and PGD) are used to evaluate the models, with results reported in terms of Attack Success Rate (ASR). The best result is highlighted in bold, while the second-best result is underlined.}

    \begin{minipage}{\textwidth}
    \centering
    \resizebox{0.7\textwidth}{!}{%
    \begin{tabular}{lccccccc}
        \toprule
        & \multicolumn{6}{c}{Epigenetic Marks Prediction} \\
        \cmidrule(lr){2-7}
        Attack & H3 & H3K14ac & H3K36me3 & H3K4me1 & H3K4me2 & H3K4me3 \\
        \midrule
        TextFooler  & \underline{65.28} & \textbf{100.00} & \textbf{100.00} & \textbf{100.00} & \textbf{100.00} & \textbf{100.00} \\
        PGD         & 29.13 & 23.43 & 21.88 & 29.53 & 31.75 & 22.90 \\
        BERT\_Attack & \textbf{100.00} & \textbf{100.00} & \underline{99.84} & \textbf{100.00} & \underline{95.67} & \textbf{100.00} \\
        \bottomrule
    \end{tabular}
    }
    \end{minipage}

    \begin{minipage}{\textwidth}
    \centering
    \resizebox{0.7\textwidth}{!}{%
    \begin{tabular}{lccccccc}
        \toprule
        & \multicolumn{4}{c}{Epigenetic Marks Prediction} & \multicolumn{3}{c}{Promoter Detection (300bp)} \\
        \cmidrule(lr){2-5} \cmidrule(lr){6-8}
        Attack & H3K79me3 & H3K9ac & H4 & H4ac & all & notata & tata \\
        \midrule
        TextFooler  & \textbf{100.00} & \textbf{100.00} & \underline{63.67} & \textbf{100.00} & \underline{53.67} & \textbf{100.00} & \underline{24.35} \\
        PGD         & 24.51 & 26.87 & 28.29 & 22.67 & 29.39 & 2.19 & 13.01 \\
        BERT\_Attack & \textbf{100.00} & \textbf{100.00} & \textbf{91.56} & \textbf{100.00} & \textbf{100.00} & \textbf{100.00} & \textbf{100.00} \\
        \bottomrule
    \end{tabular}
    }
    \end{minipage}

    \begin{minipage}{\textwidth}
    \centering
    \resizebox{0.7\textwidth}{!}{%
    \begin{tabular}{lcccccccccc}
        \toprule
        & \multicolumn{5}{c}{Transcription Factor Prediction (Human)} & \multicolumn{3}{c}{Core Promoter Detection} \\
        \cmidrule(lr){2-6} \cmidrule(lr){7-9}
        Attack & tf0 & tf1 & tf2 & tf3 & tf4 & all & notata & tata \\
        \midrule
        TextFooler  & \textbf{100.00} & \textbf{100.00} & \textbf{100.00} & \textbf{100.00} & \textbf{100.00} & \textbf{100.00} & \textbf{100.00} & 24.50 \\
        PGD         & 22.17 & 21.76 & 26.96 & 23.33 & 26.32 & 45.80 & 28.48 & \underline{28.69} \\
        BERT\_Attack & \underline{99.81} & \textbf{100.00} & \underline{98.91} & \textbf{100.00} & \textbf{100.00} & \textbf{100.00} & \textbf{100.00} & \textbf{100.00} \\
        \bottomrule
    \end{tabular}
    }
    \end{minipage}

    \begin{minipage}{\textwidth}
    \centering
    \resizebox{0.5\textwidth}{!}{%
    \begin{tabular}{lccccc}
        \toprule
        & \multicolumn{5}{c}{Transcription Factor Prediction (Mouse)} \\
        \cmidrule(lr){2-6}
        Attack & 0 & 1 & 2 & 3 & 4 \\
        \midrule
        TextFooler  & \underline{31.09} & \textbf{100.00} & 13.31 & 8.88 & \underline{80.71} \\
        PGD         & 9.09 & 28.69 & \underline{13.56} & \underline{26.81} & 28.02 \\
        BERT\_Attack & \textbf{100.00} & \underline{98.99} & \textbf{100.00} & \textbf{100.00} & \textbf{100.00} \\
        \bottomrule
    \end{tabular}
    }
    \end{minipage}
\end{table}

\clearpage
\def\arxivfont{\rm}
\bibliographystyle{plainnat}

\if
\baselineskip=.98\baselineskip
\let\originalthebibliography\thebibliography
\renewcommand\thebibliography[1]{
  \originalthebibliography{#1}
  \setlength{\itemsep}{0pt plus 0.3ex}
}
\fi

\bibliography{refs}

\end{document}